\documentclass{ieeeaccess}
\usepackage{cite}
\usepackage{amsmath,amssymb,amsfonts}
\usepackage{graphicx}
\usepackage{textcomp}
\usepackage{algorithm}
\usepackage{algorithmicx}%
\usepackage{algpseudocode}%
\usepackage{hyperref}
\usepackage{cleveref}
\usepackage{multirow}
\usepackage{amsmath,amssymb,amsfonts}%
\usepackage{amsthm}%
\usepackage{mathrsfs}%
\usepackage{booktabs}%
\usepackage{subcaption}
\usepackage{url} 
\usepackage{listings}%

\usepackage{bm}
\makeatletter
\AtBeginDocument{\DeclareMathVersion{bold}
\SetSymbolFont{operators}{bold}{T1}{times}{b}{n}
\SetSymbolFont{NewLetters}{bold}{T1}{times}{b}{it}
\SetMathAlphabet{\mathrm}{bold}{T1}{times}{b}{n}
\SetMathAlphabet{\mathit}{bold}{T1}{times}{b}{it}
\SetMathAlphabet{\mathbf}{bold}{T1}{times}{b}{n}
\SetMathAlphabet{\mathtt}{bold}{OT1}{pcr}{b}{n}
\SetSymbolFont{symbols}{bold}{OMS}{cmsy}{b}{n}
\renewcommand\boldmath{\@nomath\boldmath\mathversion{bold}}}
\makeatother

\def\BibTeX{{\rm B\kern-.05em{\sc i\kern-.025em b}\kern-.08em
    T\kern-.1667em\lower.7ex\hbox{E}\kern-.125emX}}

\definecolor{bostonuniversityred}{rgb}{0.7, 0.0, 0.0}
\definecolor{comments}{RGB}{0,0,150}

\definecolor{gray}{cmyk}{0,0,0,.8}

\begin{document}
\history{Date of publication xxxx 00, 0000, date of current version xxxx 00, 0000.}
\doi{10.1109/ACCESS.2024.0429000}

\title{Personalised Insulin Adjustment with Reinforcement Learning: An In-Silico Validation for People with Diabetes on Intensive Insulin Treatment}
\author{\uppercase{Maria Panagiotou}\authorrefmark{1,2},
Lorenzo Brigato\authorrefmark{1}, 
Vivien Streit\authorrefmark{3}, 
Amanda Hayoz\authorrefmark{3}, 
Stephan Proennecke\authorrefmark{3}, 
Stavros Athanasopoulos\authorrefmark{4},
Mikkel T. Olsen\authorrefmark{5},
Elizabeth J. den Brok\authorrefmark{6},
Cecilie H. Svensson\authorrefmark{5},
Konstantinos Makrilakis\authorrefmark{4},
Maria Xatzipsalti\authorrefmark{7},
Andriani Vazeou\authorrefmark{7},
Peter R. Mertens\authorrefmark{8},
Ulrik Pedersen-Bjergaard\authorrefmark{5,9},
Bastiaan E. de Galan\authorrefmark{6,10,11},
Stavroula Mougiakakou\authorrefmark{1}, and
on behalf of the MELISSA consortium
}

\address[1]{ARTORG Center, University of Bern, Bern, Switzerland}
\address[2]{Graduate School for Cellular and Biomedical Sciences, University of
Bern, Switzerland}
\address[3]{Debiotech SA, Lausanne, Switzerland}
\address[4]{Diabetes Center, National and Kapodistrian University of Athens,
Athens, Greece}
\address[5]{Department of Endocrinology and Nephrology, Nordsjællands Hospital,
Hillerød, Denmark}
\address[6]{CARIM School for Cardiovascular Diseases, Maastricht University,
Maastricht, The Netherlands}
\address[7]{P \& A Kyriakou’ Children’s Hospital, Athens, Greece}
\address[8]{Department of Kidney and Hypertension Diseases, Diabetology and
Endocrinology, Otto-Von-Guericke-Univeristat Magdeburg, Magdeburg,
Germany}
\address[9]{Department of Clinical Medicine, Faculty of Health and Medical Sciences, University of Copenhagen, Copenhagen, Denmark}
\address[10]{Department of Internal Medicine, Maastricht University Medical
Centre+, Maastricht, The Netherlands}
\address[11]{Department of Internal Medicine, Radboud University Medical
Centre, Nijmegen, The Netherlands}
\tfootnote{This work has been funded by the European Commission and the Swiss
Confederation-State Secretariat for Education, Research and Innovation (SERI)
within the project 101057730 Mobile Artificial Intelligence Solution for Diabetes
Adaptive Care (MELISSA)}

\markboth
{Panagiotou \headeretal: Preparation of Papers for IEEE TRANSACTIONS and JOURNALS}
{Panagiotou \headeretal: Preparation of Papers for IEEE TRANSACTIONS and JOURNALS}

\corresp{Corresponding author: Stavroula Mougiakakou (e-mail: stavroula.mougiakakou@unibe.ch).}

\begin{abstract}
Despite recent advances in insulin preparations and technology, adjusting insulin remains an ongoing challenge for the majority of people with type 1 diabetes (T1D) and longstanding type 2 diabetes (T2D). 
In this study, we propose the Adaptive Basal-Bolus Advisor (ABBA), a personalised insulin treatment recommendation approach based on reinforcement learning for individuals with T1D and T2D, performing self-monitoring blood glucose measurements and multiple daily insulin injection therapy. 
We developed and evaluated the ability of ABBA to achieve better time-in-range (TIR) for individuals with T1D and T2D, compared to a standard basal-bolus advisor (BBA). 
The in-silico test was performed using an FDA-accepted population, including 101 simulated adults with T1D and 101 with T2D. 
An in-silico evaluation shows that ABBA significantly improved TIR and significantly reduced both times below- and above-range, compared to BBA. ABBA's performance continued to improve over two months, whereas BBA exhibited only modest changes. 
This personalised method for adjusting insulin has the potential to further optimise glycaemic control and support people with T1D and T2D in their daily self-management. 
Our results warrant ABBA to be trialed for the first time in humans.
\end{abstract}

\begin{keywords}
Adaptive system, diabetes, personalisation, reinforcement learning
\end{keywords}

\titlepgskip=-21pt

\maketitle

\section{Introduction}
\label{sec:introduction}
Management of type 1 diabetes (T1D) or type 2 diabetes (T2D) in the insulin-deficient state with therapeutic insulin aims to mimic normal physiology. 
For people with diabetes on multiple daily injection (MDI) insulin therapy, this requires the combination of a long-acting insulin (analogue) to cover the continuous (i.e. basal) release of insulin in fasting states with short-acting insulin (i.e. bolus/prandial insulin) to mimic the release of insulin in response to meals and elevated glucose levels. 
Self-monitoring of blood glucose (SMBG) as well as estimating carbohydrate (CHO) content of meals are, among other parameters, pivotal in fine-tuning the doses (both basal and prandial) of insulin. 
Recent technological advances, including continuous glucose monitoring (CGM), insulin pumps, and (hybrid) closed-loop insulin delivery systems, can potentially alleviate the burden of daily diabetes management.

Hybrid closed-loop systems, such as Tandem Control-IQ \cite{breton2021one}, CamAPS FX \cite{nwokolo2023camaps}, Omnipod 5 \cite{berget2022clinical}, Medtronic MiniMed™ 780G \cite{silva2022real}, have demonstrated significant improvements in various aspects of glycaemic control, including time-in-range, reduction of hypoglycaemia, and HbA1c reduction for adults, children, and young people with type 1 diabetes \cite{lameijer2023real, eldib2024evaluation, peacock2023systematic}. 
Importantly, hybrid closed-loop systems effectively manage both intra- and inter-individual variability through feedback control mechanisms that adjust insulin delivery in response to glucose fluctuations, parametrisation options like customisable basal profiles, carbohydrate ratios, insulin sensitivity factors, and target glucose levels. 
For example, the Medtronic MiniMed™ 780G allows personalised glucose targets and delivers automatic basal insulin and correction boluses every 5 minutes, providing greater customisation in insulin therapy and accommodating daily glucose variability without user intervention \cite{silva2022real}. 
The Tandem Control-IQ predicts glucose levels 30 minutes ahead and adjusts insulin delivery accordingly, with dedicated sleep and exercise modes \cite{breton2021one}. 
Additionally, the Omnipod 5 system uses a total daily insulin (TDI) delivery strategy for insulin automation, offering adjustable glucose targets for different times of the day. It also includes an automated Mode with a static basal rate that does not adjust based on CGM values and an activity Mode, which can be set for 1 to 24 hours with a glucose target of 150 mg/dL and reduced insulin delivery
\cite{berget2022clinical}. 
CamAPS FX employs an adaptive model predictive control algorithm to adjust insulin infusion every 8–12 minutes, optimising dosing based on glucose trends. The algorithm, initialised using participant weight and TDI dose, gradually adapts based on glucose patterns and insulin action duration to enhance compatibility with faster-acting insulins \cite{nwokolo2023camaps}.

Building on these advancements in hybrid closed-loop systems, recent research has explored the potential of reinforcement learning (RL), a subset of artificial intelligence (AI), to further enhance personalised insulin delivery for people with T1D. 
RL-based approaches have been proposed within both fully closed-loop \cite{hettiarachchi4226648g2op2c, fox2020deep, louis2022safe} and hybrid closed-loop systems \cite{jafar2021long, zhu2020basal, zhu2020insulin, sun2018dual}.
Pioneering work by Daskalaki et al. \cite{daskalaki2013personalized} proposed an actor-critic (AC) method initialised with information transferred from insulin to glucose signals. 
Sun et al. \cite{sun2018dual} extended and further validated this algorithm, while with the same algorithmic approach, the authors introduced the use of SMBG measurements and MDI therapy instead of CGM data and pump therapy \cite{sun2019reinforcement}.
More recently, studies have increasingly integrated CGMs into advanced algorithms for managing T1D. 
For instance, a modular deep RL algorithm based on the proximal policy optimisation algorithm was designed to fully automate glucose control, utilising CGMs \cite{hettiarachchi4226648g2op2c}. 
Similarly, Zhu et al. \cite{zhu2020basal} developed a deep-Q learning agent (DQN) to predict basal insulin values and a deep deterministic policy gradient AC model for insulin bolus control, both supported by CGMs \cite{zhu2020insulin}. 
Another study by Jafar et al. \cite{jafar2021long} proposed a Q-learning approach to adaptively optimise CHO ratios and basal rates, leveraging CGMs for continuous feedback.
Fox et al. \cite{fox2020deep} employed the soft AC algorithm to develop glucose control policies for closed-loop blood glucose control, with CGMs. 
Furthermore, recent work combined evolutionary, DQN, AC, and uncertainty estimation algorithms to adjust insulin sensitivity and carbohydrate-to-insulin ratios for meal boluses and reference basal rates in pump therapy and insulin pen usage, all facilitated by the continuous data from CGMs \cite{louis2022safe}.
In addition, Yoo et al. \cite{yoo2024intelligent} introduced a Dual Basal-Bolus agent, by combining offline RL with an online fine-tuning technique for individuals with type 1 diabetes using MDI and CGMs.

While much attention has been given to therapeutic advancements for people with T1D, there is an increasing recognition that insulin-treated individuals with T2D encounter similar challenges in meeting recommended glycaemic targets \cite{daly2021technology, wang2023optimized}.
The most recent study proposed a model-based RL framework to determine the optimal insulin regimen by evaluating rewards associated with the glycaemic state through interactions with person's models. 
The study demonstrates that the algorithm outperformed other methods in insulin titration optimisation and showed promising results in a  blinded-feasibility trial \cite{wang2023optimized}.
Additionally, Liu et al. \cite{liu2020deep} developed an algorithm based on the DQN to recommend oral antidiabetic drugs and insulin. 
The evaluation of the algorithm involved assessing its prescriptions' concordance with recommendations and comparing clinical outcomes, showing significantly better long-term outcomes and reduced hypoglycaemia events compared to traditional methods. 
Overall, the DQN model demonstrates effective learning of prescription patterns, which leads to improved diabetes health outcomes.

Despite these advancements, hybrid closed-loop systems rely heavily on continuous sensor data and are not universally accessible due to high costs and regulatory constraints. 
Moreover, there is a substantial group of individuals with diabetes who, for various reasons, are not treated with insulin pumps with automated insulin delivery but instead remain on insulin pens and SMBG. 
This highlights the unmet need for alternative approaches that can provide personalised insulin recommendations for this population.
This study introduces an Adaptive Basal-Bolus Advisor, ABBA, an extended version of previous works \cite{sun2019reinforcement, sun2018dual}, designed specifically for individuals with diabetes using SMBG and MDI therapy (Supplementary Methods).  
The AI-based treatment decision system for adjusting basal and bolus insulin for both T1D and T2D on MDI is designed to be reliable, unbiased, cost-effective, and operates, except for a short initialisation period (two weeks), independently of continuous glucose monitoring.

\section{Methods}
The first key addition compared to the previous ABBA version \cite{sun2019reinforcement} includes another set of agents controlling the introduced physiologic state (PS) factor.
Secondly, we added the insulin-on-board (IOB) as part of the standard bolus calculator (BC) formula \cite{schmidt2014bolus}.
Furthermore, we use an adaptive optimiser which improves training stability.
We introduced the hyperparameter selection procedure based on the two weeks' CGM data collected to further improve robustness across different people with diabetes and sustain a more extreme but realistic simulation scenario. 
Finally, we substitute the gaussian policy implemented in the previous version with a linear deterministic policy, which resulted in better stability and simplified the overall algorithm complexity.
Lastly, our proposed ABBA supports insulin-treated people with T2D due to the adaptation of the cost function and hyperparameters compared to previous works \cite{sun2018dual, sun2019reinforcement, daskalaki2016model}.

\subsection{ABBA pipeline}

\begin{figure*}[!tb]
	\centering
 	\includegraphics[width=0.8\textwidth]{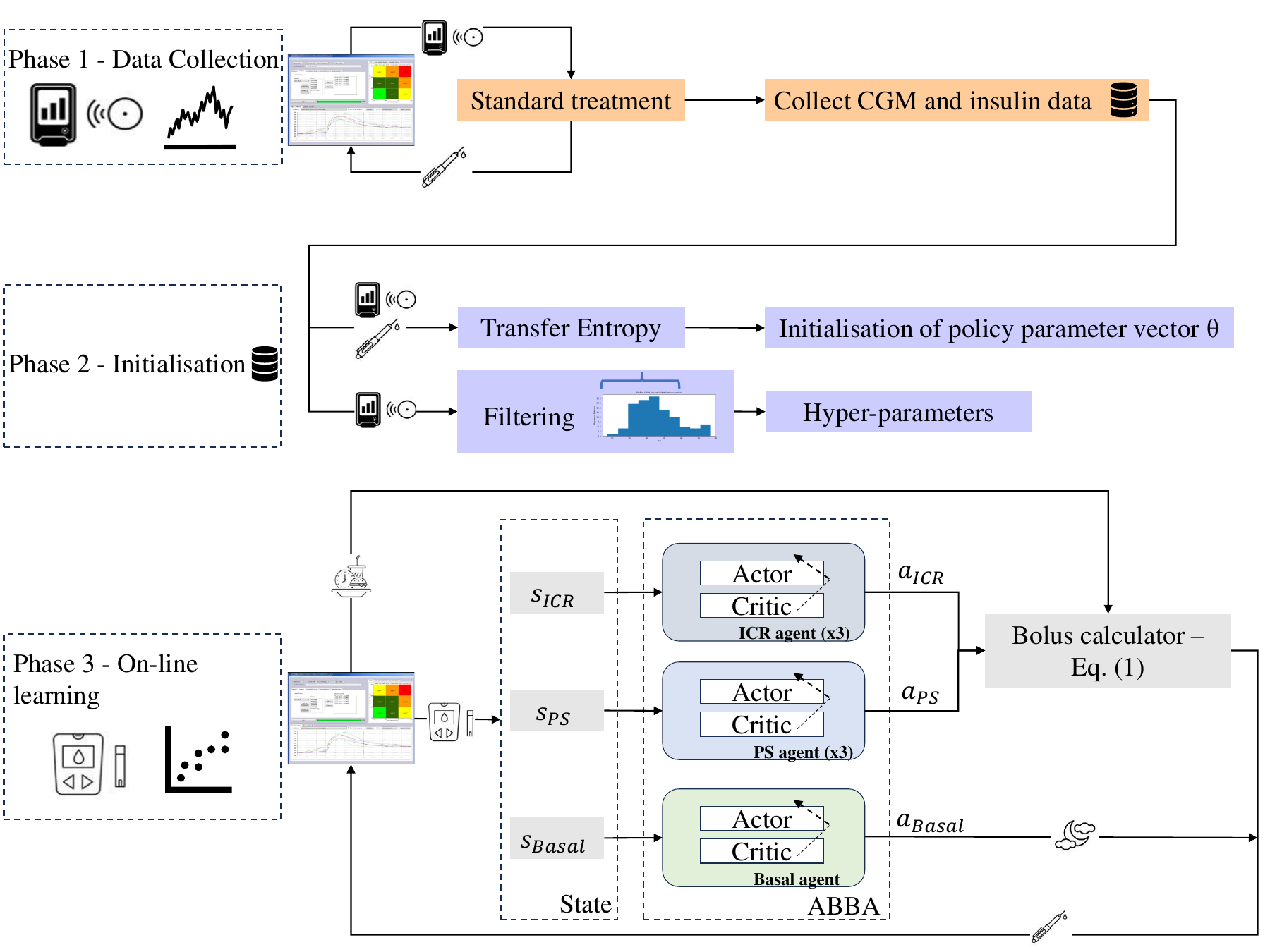}
	\caption{Outline of the ABBA pipeline, consisting of three phases: Data Collection, Initialisation, and On-line Learning. Data Collection involves gathering standard treatment data over two weeks. Initialisation uses this data to set up ABBA's parameters. On-line Learning adapts insulin doses in real-time based on glucose measurements and carbohydrate intake.}
	\label{fig:AbbaOutline}
\end{figure*}

\Cref{fig:AbbaOutline}, depicts the proposed pipeline employed for ABBA.
The pipeline is split into three consecutive phases, namely Data Collection, Initialisation, and On-line Learning.

\subsubsection{Phase 1: Data Collection}

The first phase is the data collection, which lasts for two weeks. 
In this phase, the person employs a standard treatment, following insulin suggestion from a standard BBA, based on insulin-to-carbohydrate ratio (ICR), correction factor (CF), and basal insulin provided by the simulator for T1D or by standard equations for T2D (Supplementary Data Collection). \footnote{In real-world clinical trials, these values are provided by the treating physician.}
During the first phase, CGM measurements and insulin data were collected.

\subsubsection{Phase 2: Initialisation} \label{subsec:init}

The second phase involves the initialisation of the algorithm, in which the collected data is used to initialise the policy parameter vector and to select the algorithm hyperparameters, based on the CGM data.
More details of the initialisation phase are broadly explained in the Supplementary Initialisation, and for results showing the effect of the initialisation, see the Supplementary Initialisation Period section.

\subsubsection{Phase 3: On-line Learning}

The third phase consists of the actual learning process.
ABBA is based on the actor-critic RL algorithm and is characterised by two complementary agent parts: the critic, responsible for evaluating the control policy, and the actor, responsible for improving the control policy \cite{konda1999actor}. 
ABBA suggests personalised basal and bolus insulin doses. 
Information on the current blood glucose and meal CHO content are needed to calculate the bolus insulin.
To calculate the bolus insulin recommendation, an individual's ICR and the PS is adapted and used in the following bolus calculator ($BC_{PS}$) formula:

\begin{equation}
	BC_{PS} = (\frac{CHO}{ICR} + \frac{G_c - G_T}{CF}) \cdot PS - IOB
    \label{eq:bc1}
\end{equation}

The PS is subject to large intra- and inter-individual variation, and since there is no research supporting quantification of the proportion, people with diabetes are left to a trial-and-error approach \cite{schmidt2014bolus}. 
Differently, ABBA adapts the PS factor based on the individual's glycaemic profile changes. 
In practice, PS is determined based on post-prandial glucose levels from the corresponding time on the previous day.
Finally, if insulin from previously administered boluses is still active, IOB is subtracted from the suggested bolus insulin dose.

The algorithm updates the ICR and the PS value after each corresponding meal.
The basal insulin is adjusted before its injection by performing a real-time prediction. 
Three different agents for both the ICR and PS are determined: one for breakfast hours, one for lunch hours, and one for dinner hours.

\subsubsection{System State}
We first define the notation to address the different timings involved in the algorithm update process and modelling.
ABBA continuously updates over a set of $K = |\mathcal{D}|$ days with each day indexed with $k$ ($k \in \mathcal{D} = \{1, \hdots, K\}$).
The basal agent is updated once per day.
On each day $k$, we update the bolus agents $T_{k}=|\mathcal{M}|$ times with each update $t \in \mathcal{M} = \{t_{k}, t_{k} + 1, \hdots, T_{k}\}$.
Note that the total number of meal updates $T_{k}$ may vary across days, e.g., skipping breakfast or having multiple snacks is allowed.
In this work, we fix $T_{k} = 3$, one for each main meal.
We finally index blood glucose (BG) measurements recorded for an update $t$ as $\tau \in \mathcal{B} = \{\tau_{t_{k}}, \tau_{t_{k}} + 1, \hdots, S_{t_{k}}\}$ and for an update $k$ as $\upsilon \in \mathcal{C} = \{\upsilon_{k}, \upsilon_{k} + 1, \hdots, R_{k}\}$.
The number of measurements per update may vary.
For example, on one afternoon, the person might document three readings (one after a meal and two additional ones), whereas on another afternoon, only the post-prandial measurement may be recorded.   
In this manner, ABBA seemingly adapts to a variable number of measurements per step.

Now, let us define the glucose error $G^{\Delta}$ with respect to a measurement as:

\begin{equation}
    G^{\Delta} =
    \begin{cases}
        G - G_h & \text{if } G > G_h \\
        G - G_l & \text{if } G < G_l \\
        0 & \text{otherwise}
    \end{cases}
\end{equation}

with $G_h = 180$mg/dL and $G_l = 70$mg/dL are respectively the hyper- and hypoglycaemia bounds.
The building block of the ABBA system state and cost is the feature vector $\mathbf{F} \in \mathbb{R}^{2}$.
For the basal agent, $\mathbf{F}_{k}$ is computed as:

\begin{equation}
\mathbf{F}_{k} = \left[
\frac{1}{N_{h}} \left(\sum_{\upsilon} G^{\Delta}_{\upsilon} + \sum_{t,\tau} G^{\Delta}_{t,\tau}\right),
-\frac{1}{N_{l}} \left(\sum_{\upsilon} G^{\Delta}_{\upsilon} + \sum_{t,\tau} G^{\Delta}_{t,\tau}\right)
\right]
\label{eq:features_basal}
\end{equation}
meaning that all recorded measurements over a day, including the bolus recordings, are considered.
For the bolus agents the features $\mathbf{F}_{t}$ are computed as:

\begin{equation}
\mathbf{F}_{t} = \left[
\frac{1}{N_{h}} \sum_{\tau} G^{\Delta}_{\tau},
-\frac{1}{N_{l}} \sum_{\tau} G^{\Delta}_{\tau}
\right]
\label{eq:features_bolus}
\end{equation}

For both cases, $N_{h}$ is the number of samples above the hyperglycaemia and $N_{l}$ below the hypoglycaemia thresholds.
We \textit{min-max} normalise the features to $[0, 1]$ range.

A graphical example of how the features are arranged and computed throughout the day is provided in Supplementary Figure 1.
We design different states for each group of agents to provide more relevant information and improve the prediction ability.
The Basal agent implements the state $\mathbf{s}^{bas}_{k} \in \mathbb{R}^{4}$ as:

\begin{equation}
\mathbf{s}^{bas}_{k} = \left[\mathbf{F}_{k}, \mathbf{b}_{k}
 \right]
\label{eq:s_basal}
\end{equation}
with $\mathbf{b}_{k}$ containing features quantifying the difference in blood glucose between morning and night measurements.
Simply speaking, the components of $\mathbf{b}_{k}$ are different from zero if overnight a hypo- or hyperglycaemic event has occurred.
More rigorously, $\mathbf{b}_{k} \in \mathbb{R}^{2}$ is defined as:

\begin{equation}
\mathbf{b}_{k} = 
\left[
\begin{aligned}
    &\left\{
    \begin{aligned}
        \multirow{2}{*}{$G_{\tau=\tau_{t_{k}}} - G_{\upsilon=R_{k-1}}$} && \text{if } G_{\tau=\tau_{t_{k}}} > G_h \\
        && \text{and } G_{\upsilon=R_{k-1}} < G_h \\
        0 && \text{otherwise}
    \end{aligned}
    \right. \\
    &\left\{
    \begin{aligned}
        \multirow{2}{*}{$G_{\upsilon=R_{k-1}} - G_{\tau=\tau_{t_{k}}}$} && \text{if }  G_{\tau=\tau_{t_{k}}} < G_{l1} \\
        && \text{and } G_{\upsilon=R_{k-1}} > G_{l1} \\
        0 && \text{otherwise}
    \end{aligned}
    \right.
\end{aligned}
\right]
\end{equation}

Note that the subscript $\tau=\tau_{t_{k}}$ indicates the first-morning measurement of the current day and $G_{l1} =90$ mg/dL.
As we described, we have three ICRs.
The state of the ICR agents with $t<T_{k}$, i.e., the breakfast and lunch agents, is:

\begin{equation}
\mathbf{s}^{{ICR}}_{t<T_{k}} = \mathbf{F}_{t}
\label{eq:s_ps}
\end{equation}

In the case of $t = T_{k}$, meaning the ICR for the dinner meal, the state is modified as follows:

\begin{equation}
\mathbf{s}^{ICR}_{t=T_{k}} = \left[\mathbf{F}_{t},\mathbf{b}_{k} \right]
\label{eq:s_icr3}
\end{equation}

We indeed also concatenate the difference between morning and night measurements.
In summary, $\mathbf{s}^{{ICR}}_{t<T_{k}} \in \mathbb{R}^{2}$ and $\mathbf{s}^{ICR}_{t=T_{k}} \in \mathbb{R}^{4}$.
For the PS agents the state is $\mathbf{s}^{{PS}}_{t} = \mathbf{F}_{t}$, $\mathbf{s}^{{PS}}_{t} \in \mathbb{R}^{2}$.

\subsubsection{Design of the Cost Function}
\begin{figure}[!ht]
\centering
\begin{subfigure}{0.8\linewidth}
    \centering
    \includegraphics[width=0.8\linewidth]{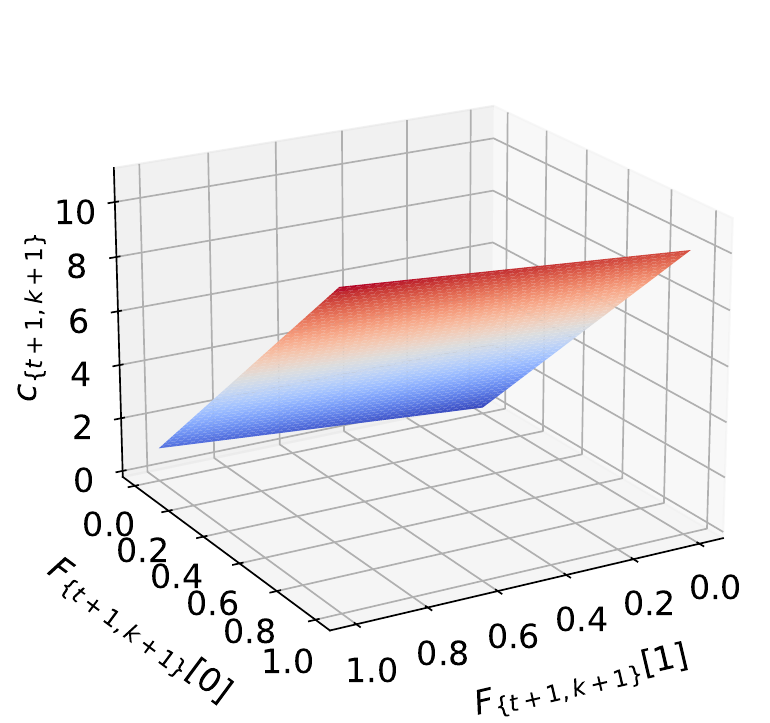}
    \caption{Cost function for T1D}
    \label{fig:costT1D}
\end{subfigure}
\vspace{0.1cm}
\begin{subfigure}{0.8\linewidth}
    \centering
    \includegraphics[width=0.8\linewidth]{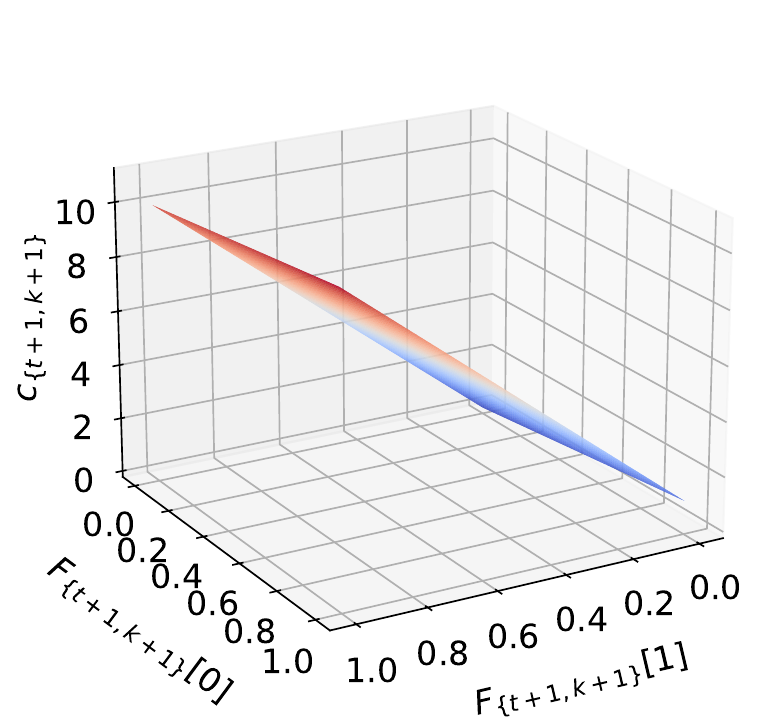}
    \caption{Cost function T2D}
    \label{fig:costT2D}
\end{subfigure}
\caption{Visualisation of the cost function (\Cref{eq: cost}).}
\label{fig:cost}
\end{figure}

ABBA minimises a modified version of the original cost function introduced by Sun et al.\cite{sun2018dual}.
Simply speaking, the cost is proportional to the features after the performed action.
The cost collected after the executed action is:

\begin{equation}
    c_{\{t+1, k+1\}} = \beta \cdot \mathbf{s}_{\{t+1,k+1\}}[0,1] = {\beta} \cdot \mathbf{F}_{\{t+1,k+1\}}^T
    \label{eq: cost}
\end{equation}

where ${\beta} = [\beta_{hypo}, \beta_{hyper}]$ is a vector containing the weights $\beta_{hypo}$ and $\beta_{hyper}$ used for scaling the hypo- and hyperglycaemia components.
For people with T1D, the aim is to minimise the hypoglycaemia events.
Thus, the weights are chosen as $\beta_{hyper} = 1$ and  $\beta_{hypo} = 10$ \cite{daskalaki2013actor}. 
On the other hand, in the case of people with T2D, the goal is to reduce the hyperglycaemia events. 
Therefore, the weights are chosen as $\beta_{hyper} = 10$ and  $\beta_{hypo} = 1$. 
\cref{fig:cost} demonstrate a visualisation of the cost function for people with T1D and T2D, respectively.

\subsubsection{Design of the Critic} 
The critic agent assesses the current control policy by estimating the long-term expected cost.
The critic component of ABBA is updated as detailed in \cite{sun2018dual}. 
It provides temporal difference (TD) error to the actor for policy optimisation, expressed as: 

\begin{equation}
d_{\{t,k\}} = c_{\{t+1,k+1\}} + \gamma \cdot V_{\mathbf{w}}(\mathbf{s}_{\{t+1,k+1\}}) - V_{\mathbf{w}}(\mathbf{s}_{\{t,k\}})
\end{equation}

where $V_{\mathbf{w}}(\mathbf{s}_{\{t,k\}})$ is the value function.
$V_{\mathbf{w}}(\mathbf{s}_{\{t,k\}})$ is parametrised as $V_{\mathbf{w}}(\mathbf{s}_{\{t,k\}}) = \mathbf{w} \cdot \mathbf{s}_{\{t,k\}}$
with $\mathbf{w}$ is the parameter vector $\mathbf{w}$.
The critic parameter vector $\mathbf{w}$ is randomly initialised at the beginning and updated according to:

\begin{equation}
    \mathbf{w}_{\{t+1,k+1\}} = \mathbf{w}_{\{t,k\}} + lr_{c} \cdot d_{\{t,k\}}\cdot \mathbf{z}_{\{t,k\}}
\end{equation}

where $lr_{c}$ denotes the learning rate and $\mathbf{z}_{\{t,k\}}$ is the eligibility vector updated as follows:

\begin{equation}
    \mathbf{z}_{\{t+1,k+1\}} = \lambda \cdot \mathbf{z}_{\{t,k\}} + \mathbf{s}_{\{t+1,k+1\}}
\end{equation}

with $\lambda$ being the eligibility trace decay factor. 
In the beginning, the eligibility vector is initialised randomly.

\subsubsection{Design of the Actor} 
The actor agent aims to optimise the control policy in order to minimise the long-term cost.
The control policy (\textit{$P_{t}$}) consists of two parts: the linear deterministic policy (\textit{$LP_{\{t,k\}}$}) and the supervisory policy (\textit{$SP_{t}$}).
Specifically, the linear deterministic policy relates the hypo- and hyperglycaemic status to the needed percentage change in the Basal, ICR, or PS for the next update. 
\textit{LP} is implemented by a linear layer parameterised by $\theta$:

\begin{equation}
    LP_{\{t,k\}} = \theta_{\{t,k\}} \cdot \mathbf{s}_{\{t,k\}}
    \label{eq:linearPolicy}
\end{equation}

The supervisory policy $SP_{t}$ provides conservative guidance to ICR agents by acting as a safety measure.
It is described as follows: 

\begin{equation}
    \resizebox{0.42\textwidth}{!}{$SP_{t} = 
        \begin{cases}
        \multirow{3}{*}{$-\alpha_{SP} \cdot \mathbf{F}_{t+1, k-1}[0]$} & \text{if } (\mathbf{F}_{t+1, k-1}[0]>0 \\
        & \text{and } \mathbf{F}_{t+1, k-1}[1] = 0) \\
        & \text{or } \mathbf{F}_{t+1, k-1}[0] > \mathbf{F}_{t+1, k-1}[1]\\
        \multirow{3}{*}{$+\alpha_{SP} \cdot  \mathbf{F}_{t+1, k-1}[1]$} & \text{if } 
        (\mathbf{F}_{t+1, k-1[1]}>0 \\
        & \text{and } \mathbf{F}_{t+1, k-1}[0] = 0) \\
        & \text{or } \mathbf{F}_{t+1, k-1}[1]>\mathbf{F}_{t+1, k-1}[0]\\
        0 & \text{if } \mathbf{F}_{t+1, k-1}[0] = \mathbf{F}_{t+1, k-1}[1] = 0
    \end{cases}$}
    \label{eq:supervisoryPolicy}
\end{equation}

with $\alpha_{SP}$ working as a hyperparameter to balance the influence of the correction. 

Therefore, the final policy is the weighted sum of the linear deterministic and supervisory policy:

\begin{equation}
    P_{t} = \alpha_{LP} \cdot LP_{t} + (1 - \alpha_{LP}) \cdot SP_{t}
    \label{eq:policy}
\end{equation}

where $\alpha_{LP}$ is defined as follows:
\begin{equation}
    \alpha_{LP} = \begin{cases}
         \multirow{6}{*}{0.5} & \text{if } ((\mathbf{F}_{t+1, k-1}[0]>0 \\
        & \text{and } \mathbf{F}_{t+1, k-1}[1] = 0) \\
        & \text{or } \mathbf{F}_{t+1, k-1}[0]>\mathbf{F}_{t+1, k-1}[1])\\
        & \text{or } ((\mathbf{F}_{t+1, k-1[1]}>0 \\
        & \text{and } \mathbf{F}_{t+1, k-1}[0] = 0) \\
        & \text{or } \mathbf{F}_{t+1, k-1}[1]>\mathbf{F}_{t+1, k-1}[0])\\
        0 & \text{if } \mathbf{F}_{t+1, k-1}[0] = \mathbf{F}_{t+1, k-1}[1] = 0 \\
        1 & \text{otherwise} \\
    \end{cases}
    \label{eq:weight}
\end{equation}

The role of $\alpha_{LP}$ is to balance the contribution of LP and SP policy to the final policy of the ICR agents. 
When the features are in range, the weighting factor $\alpha_{LP}$ is deliberately set to $0$, ensuring that the LP policy remains unweighted.
Otherwise, equal contributions to both policies are ensured when the features overcome this range.
In the case of the basal and PS agents, the final policy is equal to the linear policy.

The actor’s policy parameters $\theta$ are updated by following the policy gradient $\mathbf{g}_{\{t,k\}}$, which, having a linear policy, simply corresponds to:

\begin{equation}
    \mathbf{g}_{\{t,k\}} = \frac{d_{\{t,k\}}}{\mathbf{s}_{\{t,k\}}}
\end{equation}

Where $d_{\{t,k\}}$ is the TD error.
We employ the \texttt{Adam} optimiser \cite{kingma2014adam} with learning rate $lr_{a}$ to solve the optimisation problem. 

\subsubsection{Action} 
For the ICR and PS agents, the final action prediction is according to the following formula:

\begin{equation}
    a_t = a_{t,k-1} + m \cdot P_{t} \cdot a_{t,k-1} 
\end{equation}

For the basal agent, the action corresponds to:

\begin{equation}
    a_{k} = a_{k-1} + m \cdot P_{k} \cdot a_{k-1} 
\end{equation}

where $m$ was experimentally chosen to be $0.5$ for T1D and $1.0$ for T2D.
In this manner, ABBA recommends the basal insulin, ICR, or PS, depending on the specific agent in use.

The action space for $a_{t}$ is a continuous value in $[\frac{a_{init}}{2}, 2 \cdot a_{init}]$ where $a_{init}$ represents the initial value from the simulator or physician in real-life settings. 

To ensure safety, the basal insulin action is set to at least 25\% of the previous day's total daily dose (TDD) \cite{kuroda2011basal}. 

\begin{algorithm}[!tb]
\caption{ABBA - Pipeline}
\label{algo1}
\begin{algorithmic}[1]
\State \textcolor{bostonuniversityred}{\underline{Data Collection}}
\State \textcolor{black}{Collect 2 weeks CGM/insulin}
\State \textcolor{bostonuniversityred}{\underline{Initialisation}}
\For{\textcolor{comments}{agent $\in$ \{$Agent\_PS$, $Agent\_ICR$, $Agent\_Basal$\}}}
    \State \textcolor{black}{Initialise actor policy parameters $\theta$ using TE}
    \State \textcolor{black}{Randomly initialise critic parameter vector $\mathbf{w}$}
    \State \textcolor{black}{Filter CGM data and initialise hyperparameters}
    \State \textcolor{black}{Initialise \textcolor{black}{$a$} based on section \nameref{subsec:init}}
\EndFor

\State \textcolor{bostonuniversityred}{\underline{On-Line learning}}

\For{\textcolor{comments}{day $= k \in \mathcal{D}$}}

\State Predict basal insulin $BasI$
\State $BasI = \texttt{\textcolor{comments}{Predict}}(\mathbf{s}^{bas}_{k},~\theta^{bas}_{k},~Agent\_Basal)$
\For{\textcolor{comments}{meal step $= t \in \mathcal{M}$}}
        \State $PS = \texttt{\textcolor{comments}{Predict}}(\mathbf{s}^{PS}_{t},~\theta^{PS}_{t},~Agent\_PS)$
        \State $ICR = \texttt{\textcolor{comments}{Predict}}(\mathbf{s}^{ICR}_{t},~\theta^{ICR}_{t},~Agent\_ICR)$
        \State \textcolor{black}{Compute Bolus insulin from \Cref{eq:bc1}}
        \State \textcolor{black}{Observe $\mathbf{s}^{PS}_{t+1}$, $\mathbf{s}^{ICR}_{t+1}$, and CHO$_{t+1}$}

        \State $\theta^{PS}_{t} = \textcolor{comments}{\texttt{Update}}(\mathbf{s}^{PS}_{t+1},~\mathbf{s}^{PS}_{t},~ \theta^{PS}_{t})$
        \State $\theta^{ICR}_{t} = \textcolor{comments}{\texttt{Update}}(\mathbf{s}^{ICR}_{t+1},~ \mathbf{s}^{ICR}_{t},~\theta^{ICR}_{t})$
\EndFor
\State \textcolor{black}{Observe $\mathbf{s}^{bas}_{k+1}$}
\State $\theta^{bas}_{k+1} = \textcolor{comments}{\texttt{Update}}(\mathbf{s}^{bas}_{k+1},~ \mathbf{s}^{bas}_{k},~\theta^{bas}_{k})$
\EndFor{ \textcolor{black}{end of simulation/real-world application}}

\end{algorithmic}
\end{algorithm}

To summarise, we provide a complete overview of the algorithm in \Cref{algo1} and Algorithm 2 (Supplementary Algorithm 2).
The ABBA hyperparameters used for the experiments are provided in the Supplementary Table 2.

\subsubsection{In-silico Environment}

The experiments were run within the DMMS.Rv1.1.1© simulator which is a software designed for conducting clinical studies in virtual subjects \cite{epsilon_group_dmmsr}. 
In contrast to the  of the FDA accepted UVa/Padova Simulator \cite{man2014uva}, the newly released  DMMS.R© simulator provides additional in-silico subject population (i.e. T2D, or Pre-Diabetes). 
Furthermore, besides rapid-acting insulin, the DMMS.R© also supports simulation with long-acting insulin or oral medications, which introduces more possibilities by using different treatments for the in-silico experiments. 
Utilising this simulator, we aim to replicate real-life clinical scenarios. 
This entails involving $101$ FDA-accepted adults with T1D and $101$ adults with T2D, with an equal gender distribution, ensuring no biases related to gender (Supplementary Table 1).

\subsubsection{Experimental Scenario} \label{sec:experimental_scenario}
Our experimental scenario, including three meals and snack(s), was simulated $202$ virtual adults (101 with T1D and 101 with T2D) and was conducted for $90$ days using the DMMS.Rv1.1.1© simulator \cite{epsilon_group_dmmsr}.
For this study, a blood glucose measurement before each meal and one before bedtime each day are required, and all measurements are done with virtual SMBG devices.
A simulation day included four meals i.e., breakfast, lunch, dinner, and one snack. 
The duration of main meal was simulated for $15$ to $30$ minutes and the snack $3$ to $8$ minutes. 
\Cref{tab:scenario} reports the minimum and maximum values of the distributions from which the meal timing and carbohydrate amount were drawn.
Across the study, the meal announcement with the corresponding bolus injection took place $5$ to $15$ minutes before the meal and the basal insulin injection from 22:00 to 00:00. 
In addition, we consider that the subjects are likely to under or overestimate the carbohydrate content of meals by $70$\% and $110$\%, respectively, based on the literature \cite{brazeau2013carbohydrate, zhu2020insulin, zhu2020basal}. 
A $\pm10\%$ variability to the initial ICRs and basal insulin was added. 
Both variabilities and uncertainties follow uniform distributions.
The ``dawn phenomenon" was considered as the intra-day variability of insulin sensitivity in people with T1D. 
This refers to periodic episodes of hyperglycaemia occurring in the early morning hours before and after breakfast \cite{o2022dawn}. 
As implemented by Sun et al. \cite{sun2018dual}, in the present study, insulin sensitivity decreased daily between 04:00 and 08:00 to 50\% of its original value.
The change in insulin sensitivity happened within a timeframe of $30$ minutes. 
We assume that no virtual subjects had a no meal intake or bolus injection after they administered their basal injection. 
Finally, complete adherence to ABBA or BBA insulin recommendations was assumed.

\begin{table}[!htb]
    \centering
    \caption{Minimum and maximum values of the possible carbohydrate (CHO) amount and time of consumption for the different meals. Meal timing and CHO content of the meals were extracted from uniform distributions.}
    \label{tab:scenario}
    \begin{tabular}{|ccc|}
    \toprule
        Meal type & CHO amount [g] & Meal timing \\
        \midrule
         Breakfast & [42 - 98] & [7:00 - 9:00] \\
         \midrule
         Lunch & [60 - 140] & [12:30 - 13:30] \\
         \midrule
         Dinner & [54 - 126] & [19:00 - 20:00] \\
         \midrule
         \multirow{3}{*}{Snack} & \multirow{3}{*}{[5 - 21]} & 
         [10:00 - 11:00] or \\
         &  & [15:00 - 18:00] or \\
         &  & [21:00 - 22:30] \\
        \bottomrule
    \end{tabular}
\end{table}

A rescue meal controller, a safety measure, activated when the BG dropped below $1.7$ mmol/l ($30$ mg/dL) was added.
To restore a hypoglycaemic event, the person ingested a $20$-gram glucose tablet. 
The corresponding BG value was given to ABBA as an input.
This low cut-off threshold was intentionally chosen to observe how frequently hypoglycaemic events occurred without the participants' preventive actions.

\subsection{Metrics}

The assessment of ABBA's performance in the tested scenario was compared to the standard BBA using the following parameters 1) percentage of time in range (TIR, $3.9 - 10$ mmol/l $[70-180]$ mg/dL]), 2) percentage of time below range I (TBR I $< 3.9$ mmol/l [$70$ mg/dL]), percentage of time below range II (TBR II, $< 2.8$ mmol/l [$50$ mg/dL]), and 3) percentage of time above range (TAR, $> 10$ mmol/l [$180$ mg/dL]). 
Additionally, the frequency of hypo- and hyperglycaemia events, mean, maximum, and minimum glucose (mg/dL), HbA1c (\%), low blood glucose index (LBGI, \%), and total daily dose of insulin in units per day (U/day) were assessed.

\subsection{Statistical analysis}
The validity of the normality assumption was checked using the Lilliefors test \cite{conover1999practical}. 
The Wilcoxon signed-rank test was applied if they were skewed, otherwise the paired t-test was applied, with a significance level equal to $0.01$ to compare the differences in glycaemic parameters. 
Results are reported as mean $\pm$ std.

\section{Results}\label{sec2}
\subsection{Overall Performance}

\begin{table}[!hbt]
    \centering
    \caption{In-silico results with the full cohort over a 90-days duration. Data are mean $\pm$ standard deviation. $\dagger$ indicates statistical significance with $p < 0.01$.}
    \label{tab:results3months}
    \begin{tabular}{|lcc|}
        \toprule
         & \textbf{ABBA (n=101)} & \textbf{BBA (n=101)}\\
         \cmidrule(lr){1-3}
         \multicolumn{3}{|c|}{\textbf{Adults - T1D}} \\
         \cmidrule(lr){1-3}
         TIR, \% & $80.1 \pm 10.8^\dagger$ & $70.6 \pm 13.8$\\
         TBR I, \% & $0.9 \pm 0.9^\dagger$ & $4.4 \pm 4.8$\\
         TBR II, \% & $0.3 \pm 0.4^\dagger$ & $2.1 \pm 3.6$\\
         TAR, \% & $19.0 \pm 10.2^\dagger$ & $25.0 \pm 12.1$\\
         \# Hypoglycaemia events & $10.9 \pm 10.7^\dagger$ & $35.1 \pm 34.0$\\
         \# Hyperglycaemia events & $199.5 \pm 58.5$ & $199.4 \pm 51.3$\\
         Mean glucose, mg/dL & $148.0 \pm 9.0$ & $148.0 \pm 13.0$\\
         Maximum glucose, mg/dL & $262.0 \pm 38$ & $259.0 \pm 41.0$ \\
         Minimum glucose, mg/dL & $47.0 \pm 18.0$ & $50.0 \pm 17.0$ \\
         HbA1c, \% & $6.8 \pm 0.3$ & $6.8 \pm 0.4$ \\
         LBGI, \% & $0.27 \pm 0.22^\dagger$ & $1.28 \pm 1.38$ \\
         Total daily insulin, U/day & $43.4 \pm 11.1$ & $44.2 \pm 10.4$\\
        \cmidrule(lr){1-3}
          \multicolumn{3}{|c|}{\textbf{Adults - T2D}} \\
         \cmidrule(lr){1-3}
         TIR, \% & $80.1 \pm 16.6^\dagger$ & $68.3 \pm 19.7$ \\
         TBR I, \% & $1.0 \pm 1.4^\dagger$ & $7.9 \pm 12.6$ \\
         TBR II, \% & $0.26 \pm 0.51^\dagger$ & $3.61 \pm 7.55$ \\
         TAR, \% & $18.9 \pm 16.5^\dagger$ & $23.8 \pm 21.5$ \\
         \# Hypoglycaemia events & $11.7 \pm 19.2^\dagger$ & $52.8 \pm 98.2$ \\
         \# Hyperglycaemia events & $189.7 \pm 93.2$ & $177.0 \pm 95.7$ \\
         Mean glucose, mg/dL & $145.0 \pm 22.0$ & $145.0 \pm 37$ \\
         Maximum glucose, mg/dL & $284.0 \pm 73.0$ & $285.0 \pm 79$\\
         Minimum glucose, mg/dL & $65.0 \pm 28.0$ &  $73.0 \pm 34.0$ \\
         HbA1c, \% & $6.7 \pm 0.8$ & $6.7 \pm 1.3$ \\
         LBGI, \% & $0.35 \pm 0.38^\dagger$ & $2.05 \pm 3.30$ \\
         Total daily insulin, U/day & $56.41 \pm 20.36$ & $50.95 \pm 9.17$ \\
         \bottomrule
    \end{tabular}
\end{table}

During 90-day period, people with T1D using ABBA experienced a significantly higher TIR of $9.54 \pm 7.76 \%$, combined with a lower TBR of $3.51 \pm 4.64\%$ and a reduced TAR of $6.02 \pm 7.57\%$ when compared BBA. 
Similarly, people with T2D using ABBA showed a statistical increase in TIR by $11.80 \pm 10.76\%$, with a corresponding decrease in TBR and TAR to $6.99 \pm 12.22\%$ and $4.90 \pm 8.13\%$, respectively (\Cref{tab:results3months}).
The number of hypoglycaemia events during $90$-days period using ABBA are significantly decreased by $24.26 \pm 32.51$ and $39.80 \pm 90.46$ for individuals with T1D and T2D, respectively compared to BBA. 
The activation of the rescue meal controller occurred less often when utilising ABBA than BBA in both groups (Supplementary Figure 2).

In Supplementary Section Different Scenarios, presents experiments over three months, involving 11 individuals with type 1 diabetes and 11 with type 2 diabetes across various scenarios. ABBA consistently outperformed the standard BBA and showed continuous improvement over time.

\subsection{Performance over Time}
\begin{figure*}[!htp]
\centering
\begin{subfigure}{0.30\textwidth}
    \includegraphics[width=1.0\textwidth]{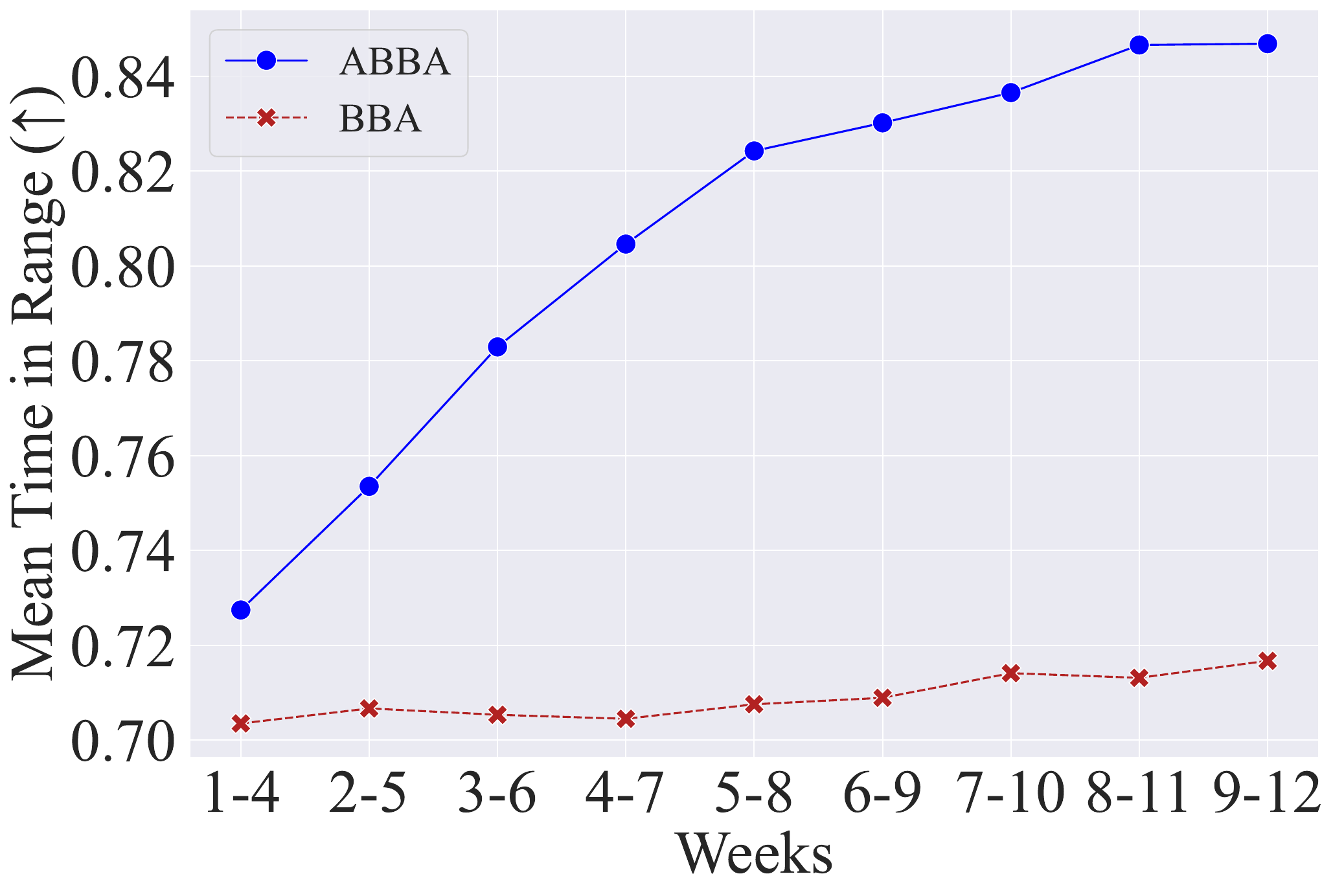}
    \caption{TIR for T1D}
    \label{fig:tir_t1d}
\end{subfigure}
\hfill
\begin{subfigure}{0.30\textwidth}
    \includegraphics[width=1.0\textwidth]{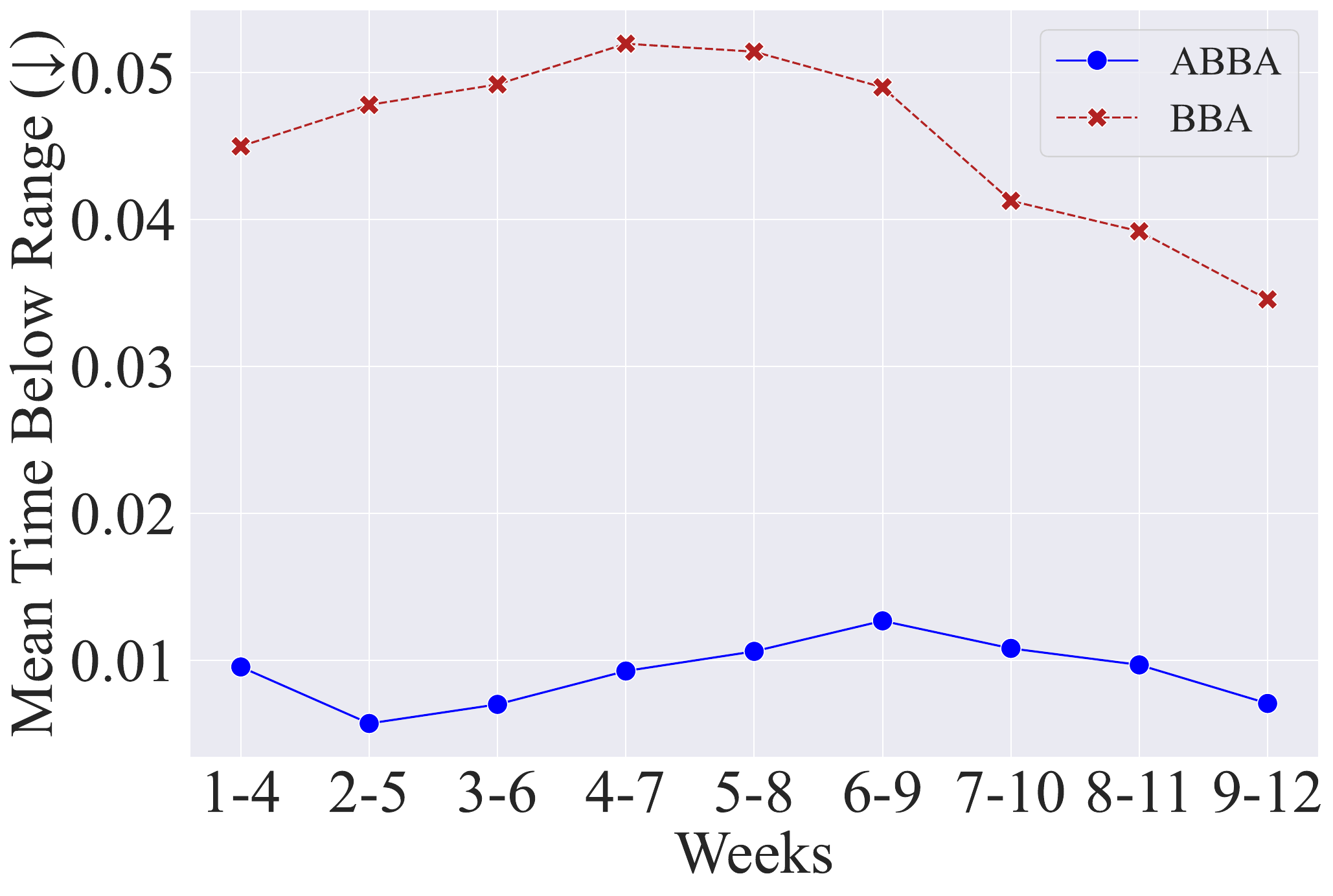}
    \caption{TBR I for T1D}
    \label{fig:tbr_t1d}
\end{subfigure}
\hfill
\begin{subfigure}{0.30\textwidth}
    \includegraphics[width=1.0\textwidth]{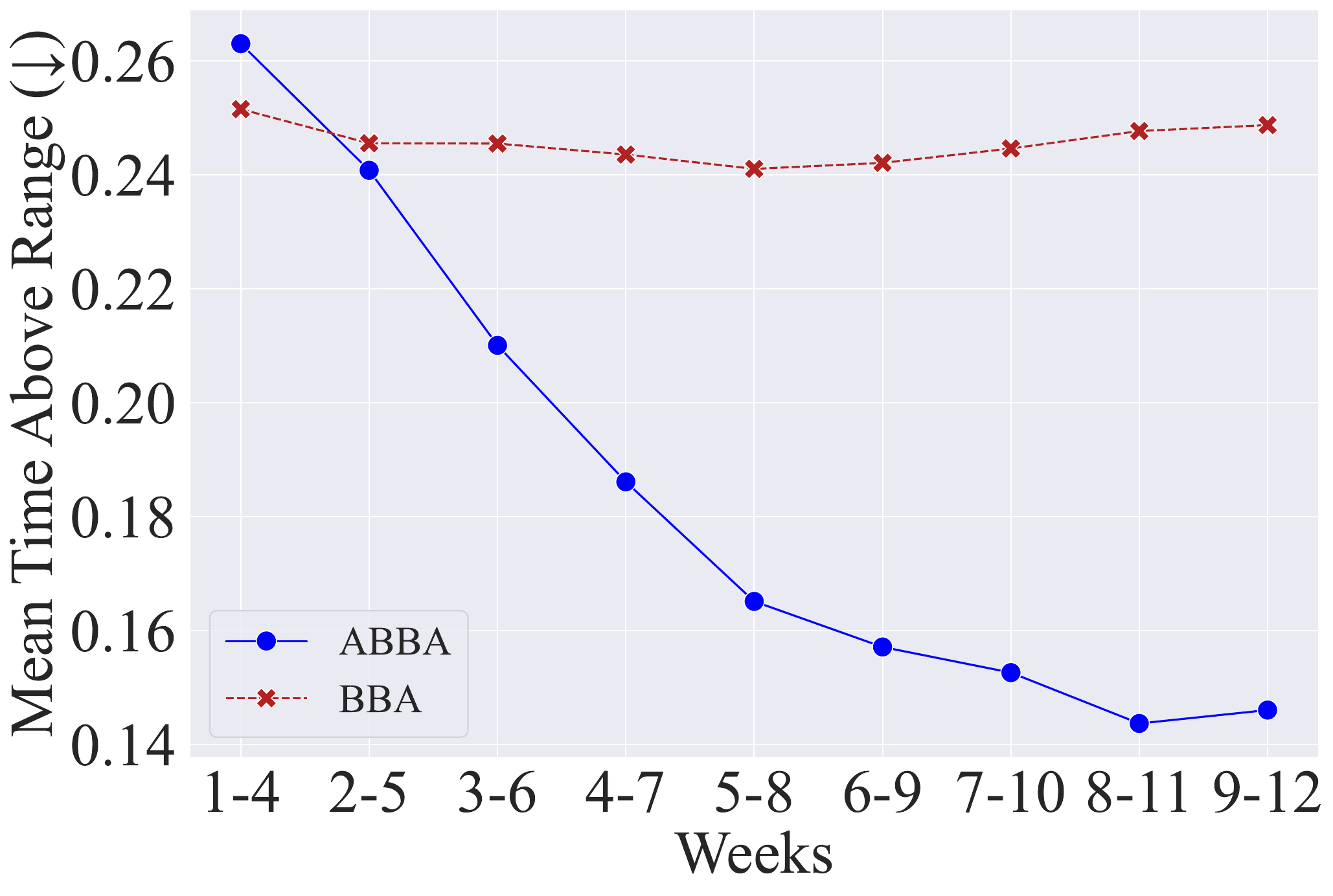}
    \caption{TAR for T1D}
    \label{fig:tar_t1d}
\end{subfigure}
\begin{subfigure}{0.30\textwidth}
    \includegraphics[width=1.0\textwidth]{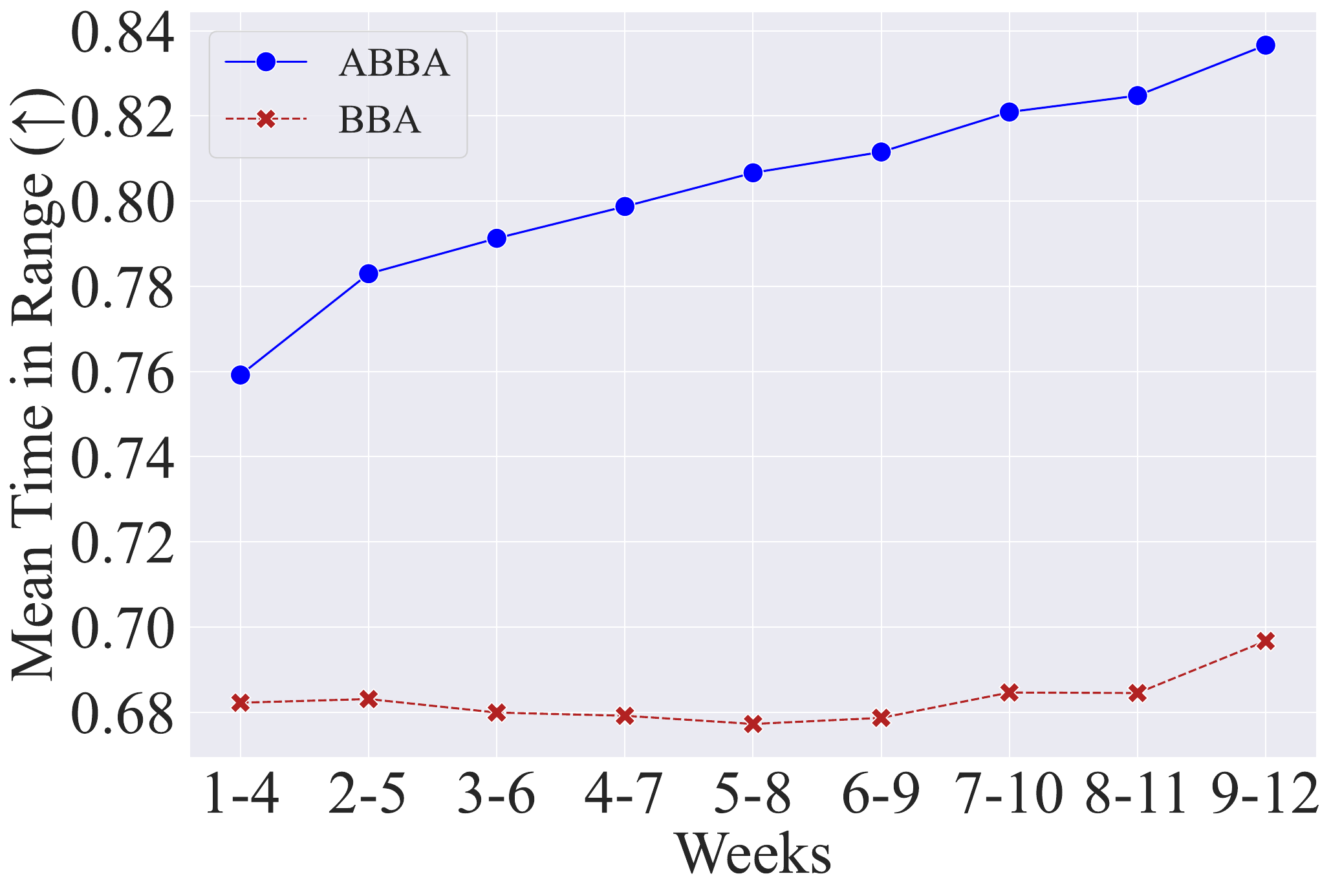}
    \caption{TIR for T2D}
    \label{fig:tir_t2d}
\end{subfigure}
\hfill
\begin{subfigure}{0.30\textwidth}
    \includegraphics[width=1.0\textwidth]{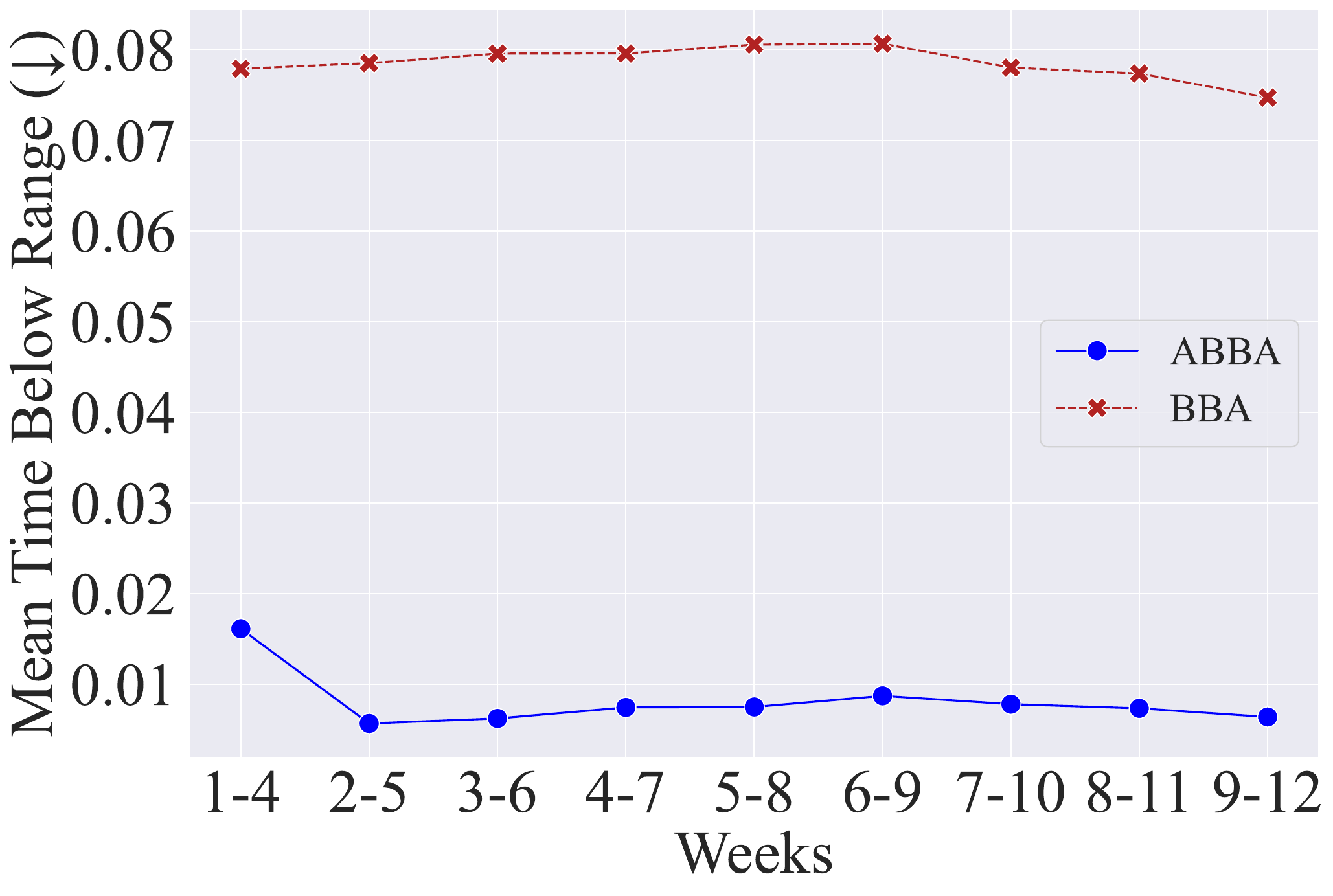}
    \caption{TBR I for T2D}
    \label{fig:tbr_t2d}
\end{subfigure}
\hfill
\begin{subfigure}{0.30\textwidth}
    \includegraphics[width=1.0\textwidth]{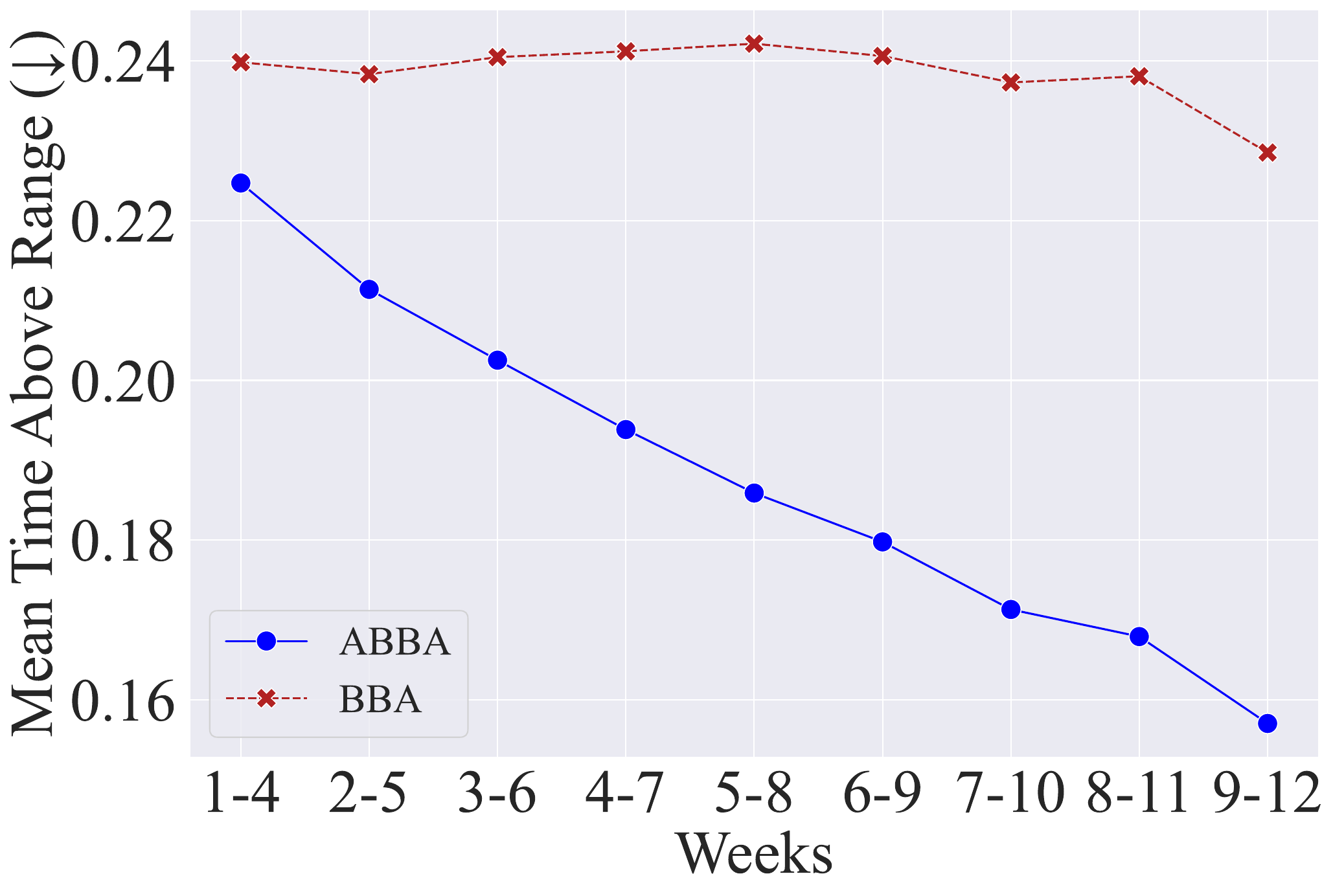}
    \caption{TAR for T2D}
    \label{fig:tar_t2d}
\end{subfigure}  
\caption{Performance of ABBA and BBA for T1D and T2D with a time window of 4 weeks shifted by steps of 1 week.}
\label{fig:figures}
\end{figure*}

During the first four weeks, ABBA demonstrated a significant increase in TIR in people with T1D by $2.39 \pm 4.32\%$ and a reduction in TBR by $3.54 \pm 4.13\%$ with no statistically significant change in TAR $(p=0.02)$, compared to BBA. 
In the last four weeks, the same regimen resulted in a significantly higher TIR of $13.01 \pm 10.41\%$ and increased reductions in TBR and TAR by $2.75 \pm 4.61\%$ and $10.27 \pm 10.25\%$, respectively, relative to BBA. 
After two months, ABBA showed a significant increase of $11.94 \pm 8.39\%$ in TIR and a decreased of $0.25 \pm 1.33\%$ and $11.69 \pm 8.46\%$ in TBR and TAR, respectively, for people with T1D. 
In contrast, BBA showed only modest improvements in TIR and TBR of $1.32 \pm 1.41\%$ and $1.04 \pm 0.95\%$, respectively with no significant change in TAR with $p>0.01$ (\Cref{fig:tir_t1d}, \Cref{fig:tbr_t1d}, and \Cref{fig:tar_t1d}, Supplementary Table 3).

Likewise, ABBA for people with T2D resulted significantly higher TIR by $7.69 \pm 9.11\%$ alongside reductions in TBR by $6.18 \pm 10.47\%$ and TAR by $1.51 \pm 5.76\%$, compared to BBA. 
In the last four weeks, people with T2D following ABBA suggestions, showed an improvement in their TIR by $13.99 \pm 11.91\%$ and lower TBR and TAR by $6.84 \pm 12.57\%$ and $7.15 \pm 10.01\%$, respectively, compared to BBA. 
After two months, ABBA led to a $7.74 \pm 5.53\%$ improvement in TIR, with corresponding reductions in TBR and TAR by $0.97 \pm 2.45\%$ and $6.77 \pm 5.89\%$, for people with T2D. 
On the other hand, BBA showed only slight improvement in TIR, TBR, and TAR of $1.45 \pm 1.47\%$, $0.32 \pm 0.91\%$, and $1.13 \pm 1.24\%$, correspondingly (\Cref{fig:tir_t2d}, \Cref{fig:tbr_t2d}, and \Cref{fig:tar_t2d}, Supplementary Table 3).

\section{Discussion}\label{sec3}
These results demonstrate a significant improvement in glycaemic control for both T1D and T2D, with increased TIR and reduced TBR and TAR when compared to BBA. 
ABBA's ability to learn and adapt over time led to continuous performance improvements, in contrast to BBA's nearly static performance, which is limited by its fixed formula.

The data-driven approach offers real-time insights and maintains low computational costs, regardless of the technology used for glucose monitoring, i.e., either SMBG, FGM, or CGM. 
Focusing on an SMBG-driven system is crucial for addressing the needs of individuals who are not treated with insulin pumps or automated insulin delivery systems and rely on MDI therapy. 
While CGM is the consensus glycaemic measurement system in the context of basal / bolus insulin therapy, it remains inaccessible for some people due to financial or regulatory barriers.
ABBA’s efficiency is evident in its minimal computational demands, requiring an average of only 0.3 milliseconds for a single prediction and update step on an AMD Ryzen 7 CPU with 16GB of RAM.  
Additionally, ABBA  undergoes continuous learning and adaptation based on real-time data inputs. 
Unlike traditional AI algorithms that rely heavily on datasets for pre-training, ABBA’s approach involves updating its knowledge and decision-making criteria dynamically as it encounters new information. 
This real-time learning process enables ABBA to adapt to changing circumstances and potentially rectify biases as they emerge. 
By avoiding pre-training on potentially biased datasets, ABBA seeks to prevent the perpetuation of biases deep-rooted in those datasets by starting with a personalised initialisation and learning purely from the data it encounters \cite{pagano2023bias}.

\begin{figure}[!bt]
    \centering
    \includegraphics[width=0.85\linewidth]{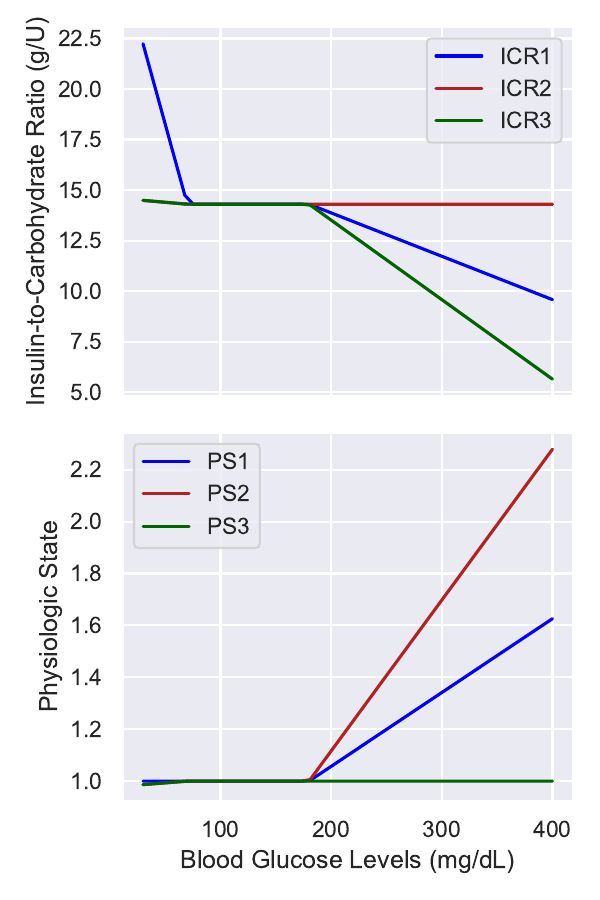}
    \caption{Visualisation of insulin-to-carbohydrate ratio (ICR) and physiologic state (PS) predictions as a function of blood glucose (BG) levels. ABBA uses a piecewise linear policy to maintain interpretability and predictability. ICR1, ICR2, and ICR3 represent the ICR values for the breakfast, lunch, and dinner agents, respectively. Similarly, PS1, PS2, and PS3 correspond to the physiologic state predictions for the breakfast, lunch, and dinner agents, respectively.}
    \label{fig:explainability}
\end{figure}

To illustrate the explainability of ABBA, we show in \Cref{fig:explainability} how the bolus agents dynamically change the ICR or PS as a function of the BG values for a random adult with T1D. 
The piece-wise linear nature of our policy makes the actions easy to predict and interpret. 
No adjustment is made when BG measurements fall within the TIR, indicating that the current ICR or PS is optimal. 
Under hypo- or hyperglycaemic conditions, the policy predicts a linear trend depending on the learned $\theta$ parameters. 
Note that having three different agents enables greater flexibility and better adaptation to the different meals.
For instance, when focusing on the hyperglycaemic range, the breakfast agent (ICR1) keeps an almost flat ICR, while the dinner agent (ICR3) significantly reduces the slope, implying that this specific person needs higher insulin dosages after dinner. 
This occurred because from 22:00 to 00:00, the person has $53\%$ TAR in the first $15$ days of our simulation.
However, as ABBA updated its parameters, this percentage dropped to $0\%$ (Supplementary Figure 3).
Furthermore, to ensure user transparency, the system shows the predicted ICR and IOB. Moreover, our algorithm uses the standard bolus calculator formula (\Cref{eq:bc1}) to estimate the bolus insulin, only with modifications to the ICR and PS factors. 
This approach, being familiar to the participants, enhances the explainability of our system.

Although the findings underscore the pivotal role of personalised treatment strategies and highlight the potential benefits of AI-based algorithmic interventions in diabetes care without the need for closed-loop systems, certain limitations should be acknowledged. 
First, the algorithm depends on the accuracy of the user input.
People with diabetes are required to precisely estimate the amount of carbohydrate content of their meal, however inaccurate carbohydrate-counting skills are common among the diabetes population. 
Exploring automated algorithms for meal nutrient content and energy estimation based on image capturing could potentially promote accuracy and streamline the process \cite{lu2020gofoodtm, papathanail2023nutritional, rahman2024food, panagiotou2023complete}.
In addition, while realistic scenarios were simulated following the literature \cite{daskalaki2013actor, daskalaki2013personalized, noaro2023personalized}, the study did not examine the effects of larger meals, which are more typical for, e.g., adolescents or scenarios where bolus injections are administered post-meal.
Additionally, our approach so far did not account for exercise, which will be included in future versions of ABBA. 
Furthermore, the lack of standardised benchmarks for meal protocols and simulators complicates consistent evaluation and comparative analysis across interventions and diabetes groups. 
This is the first analysis evaluating T2D using the DMMS.Rv1.1.1© simulator, yet direct comparisons with other studies remain challenging due to variations in cohorts, protocols, and simulator versions used.   

Up to now, randomised clinical trials on bolus advisors in people with diabetes have shown very modest improvements in HbA1c or other glycaemic outcomes and treatment satisfaction \cite{den2024effect}.
ABBA will be validated in a multinational multicenter randomised controlled clinical trial named MELISSA.\footnote{\underline{https://www.melissa-diabetes.eu/}}
The primary objective of the MELISSA trial is to demonstrate the superiority of ABBA compared to standard-of-care in people with T1D and T2D on intensive insulin treatment. In addition, participants can use an AI-based dietary assessment system to estimate meal calories and macronutrients from smartphone images \cite{papathanail2023nutritional}. To receive a bolus recommendation, participants must enter CHO before eating and inject insulin within $30$ minutes. If not, a prompt will ask for feedback on the missed injection.
Furthermore, a feasibility study is currently ongoing with the University Hospital in Geneva, Switzerland to assess the implementation of ABBA in a real-world clinical setting.
A collaborative effort has been made to develop the MELISSA application, incorporating ABBA, in compliance with the regulatory framework for medical software. 
This collaboration involves individuals with diabetes, AI experts, clinicians, and industry software developers.
This collective effort aims to bridge the gap between research and practical application, ultimately enhancing the effectiveness and accessibility of diabetes management solutions.
In future versions of ABBA, we aim to enhance the algorithm by retrospectively analysing data from the MELISSA trial.
We will focus on incorporating physical activity data and developing a more effective initialisation method that does not rely on a CGM device.

\section{Conclusion}
Overall, these in-silico results demonstrate that ABBA, driven by RL, significantly optimised TIR, TBR, and TAR in people with diabetes on intensive insulin regimen when compared to the standard BBA.
However, in-silico testing may not fully reflect real-world variability, including inconsistent adherence to insulin recommendations.
These results warrant further clinical validation of ABBA as a step towards personalised diabetes management for everyone with diabetes on intensive insulin treatment, ensuring its effectiveness and safety in real-life conditions.

\appendices

\section{Methods}
\subsection{Data Collection}
\label{subsec:data_collection}

For individuals with diabetes on multiple daily injections (MDI), it is recommended to administer long-acting insulin injections once or twice daily to control fasting blood glucose levels, along with fast-acting or bolus insulin injections before meals to mitigate postprandial glucose spikes.
For the data collection period, we follow the standard guidelines and calculate the bolus insulin with the formula of standard $BC$ \cite{schmidt2014bolus} defined as:	

\begin{equation}
    BC = \frac{CHO}{ICR} + \frac{G_c-G_T}{CF} - IOB
    \label{eq:bc2}
\end{equation}

where CHO (g) is the meal carbohydrate intake, ICR (g/U) and correction factor (CF) (mg/dL/U), respectively are the insulin-to-carbohydrates ratio and the correction factor \cite{davidson2008analysis}, $G_c$ (mg/dL) is the current blood glucose (BG) level, $G_t$ (mg/dL) is the target BG level, and IOB (U) is the insulin on board, indicating the previously injected insulin that is still acting in the body \cite{gross2003bolus}.

\subsection{Initialisation}
\label{subsec:appendix_init}

\paragraph{ICR, CF, and Basal Insulin} 
To ensure safety, the initial values for the ICR, CF, and basal insulin should be individualised. 
When implemented in clinical practice, the ICR, CF, and basal insulin of ABBA are initialised using the person's individual values as optimised by their clinician. 
A shared decision is made to fine-tune the parameters that best fits individual's needs, values and preferences.
To reflect the bias that the healthcare specialist may have when estimating “optimal” values, we randomly apply a uniform uncertainty of $\pm10\%$.
The simulator provides the default ICR, CF, and basal insulin for type 1 diabetes (T1D).
For T2D, such values are calculated using the standard equations (see paper Section Baseline Method).

\paragraph{Policy Parameters}
Policy parameters are the variables that define the behaviour of the advisor \cite{konda1999actor}.
The initialisation of the policy parameter vector $\theta$ is based on the transfer entropy (TE) between CGM data and the active insulin, as implemented by Daskalaki et al. \cite{daskalaki2016model}.
Transfer entropy measures how much information about the past behaviour of active insulin helps predict the future behaviour of insulin sensitivity pattern \cite{daskalaki2016model}.
Active insulin (AI) was estimated as the sum of IOB related to the bolus doses and basal insulin ($BasI$) infusion collected in the first two weeks:
\begin{equation}
    AI = IOB + BasI
\end{equation}

Estimation of IOB was based on the standard approach introduced by Lee et al. \cite{lee2009closed}. 
TE was calculated using the \texttt{PyInform} Python library.
The overall procedure was similarly employed in \cite{daskalaki2016model}.

\paragraph{Hyperparameters}
Hyperparameters are settings that guide how ABBA learns, such as the speed of learning or the complexity of the suggested insulin dose.
The varied responses to insulin among individuals with diabetes necessitate personalised treatment, as evidenced by discrepancies in the stability of blood glucose profiles.
Consequently, those with increased BG variability require more careful management, leading us to adopt more conservative algorithmic updates. 
We gather two weeks of BG data from $101$ simulated adults diagnosed with T1D and compute the standard deviation of BG values. 
Notably, the CGM profiles display a nearly Gaussian distribution.
Individuals with increased glycaemic variability are identified as those whose BG standard deviation is more than one standard deviation above the mean, known as a $\texttt{z-score} > 1$. 
Practically, people with T1D whose BG standard deviation exceeds $57$ are classified as people with increased glycaemic variability. 
We repeat the same approach for people with T2D. 
We define a person with T2D as having increased glycaemic variability when the BG standard deviation surpasses $59$. 
Therefore, as shown in \Cref{tab:parameters}, we decrease the AC's initial learning rates, because smaller learning rates mean slower adaptation, to support those with unstable BG profiles more effectively. 
Moreover, in case the subject in the first two weeks has BG less than $70$ mg/dL for more than $21\%$ for T1D and $42\%$ for T2D overnight, we consider them as high-risk.
The decision of this cut-off is taken with similar approach based only for the overnight CGM profiles on the first two weeks.
Following the aforementioned approach, we adjust the actor learning rate of the dinner ICR accordingly and will provide more details in appendix Section \ref{ABBA Hyperparameter Setup}.

\subsection{On-Line Learning}

\begin{figure}[tb]
    \centering
    \includegraphics[width=0.9\linewidth]{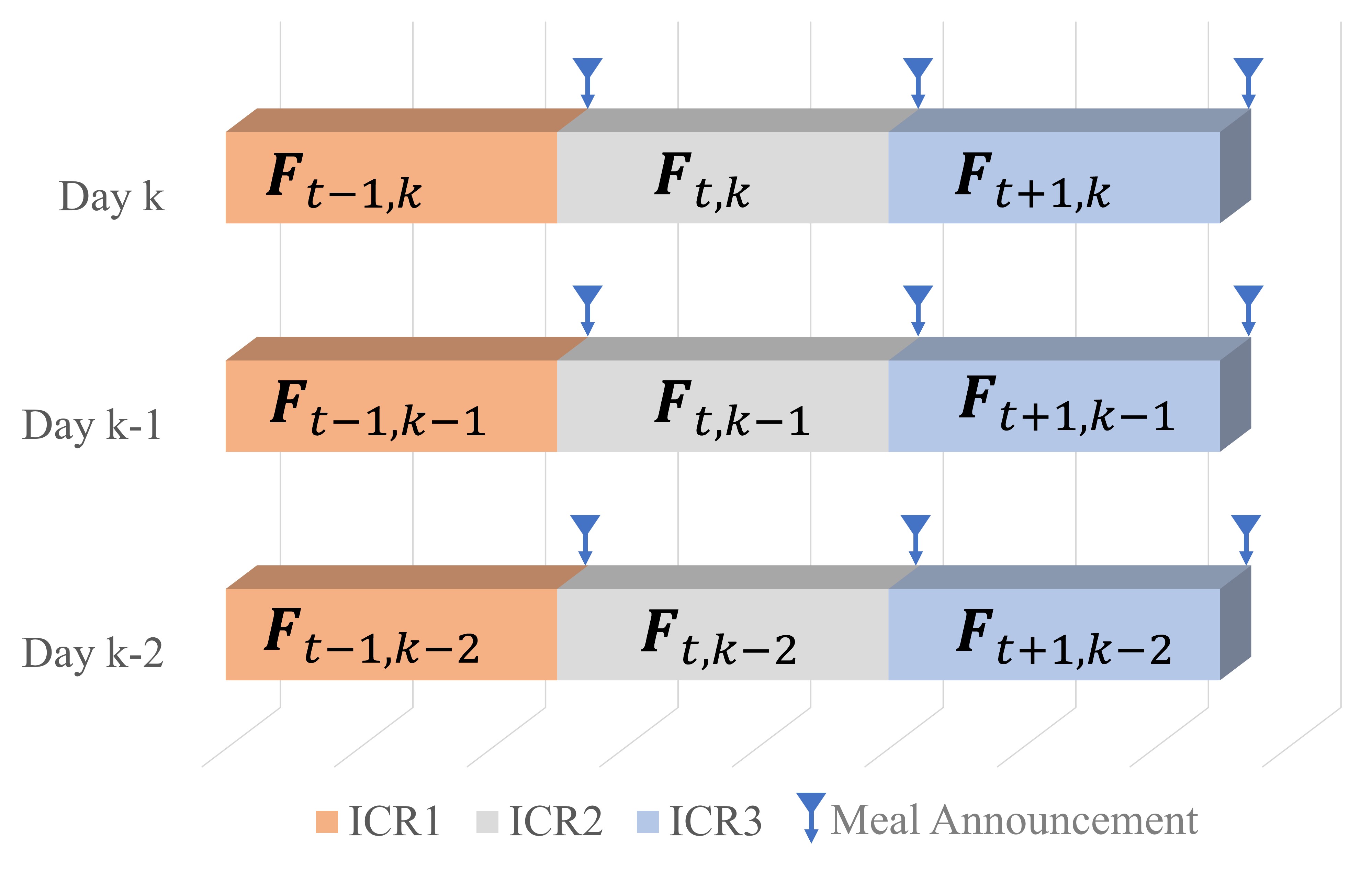}
    \caption{Simple graphical illustration of the features and corresponding ICRs and PSs aligned with the meals, assuming three meals during a day in this specific example.}
    \label{fig:features}
\end{figure}

\paragraph{Algorithm 2.} We summarise the predict and update functions of ABBA in Algorithm 2.

\setcounter{algorithm}{1}
\begin{algorithm}[ht]
\caption{\textcolor{black}{$\texttt{Update} \text{~and~} \texttt{Predict}$ functions}}
\label{algo2}
\begin{algorithmic}[2]

\Function{$\textcolor{comments}{\texttt{Update}}$}{$\mathbf{s}_{\{t+1,k+1\}}$, $\mathbf{s}_{\{t,k\}}$, $\theta_{\{t,k\}}$}

\State \textcolor{black}{Compute cost} 
\State \textcolor{gray}{$c_{\{t+1,k+1\}} = \mathbf{\beta} \cdot \mathbf{s}_{\{t+1,k+1\}}[0,1]$}
\State \textcolor{black}{Compute values from transition}
\State \textcolor{gray}{$V_{\mathbf{w}}(\mathbf{s}_{\{t+1,k+1\}}) = \mathbf{w} \cdot \mathbf{s}_{\{t+1,k+1\}}$}
\State \textcolor{gray}{$V_{\mathbf{w}}(\mathbf{s}_{\{t,k\}}) = \mathbf{w} \cdot \mathbf{s}_{\{t,k\}}$}

\State Compute TD error
\State \textcolor{gray}{$d_{\{t,k\}} = c_{\{t+1,k+1\}} + \gamma \cdot V_{\mathbf{w}}(\mathbf{s}_{\{t+1,k\}}) - V_{w}(\mathbf{s}_{\{t,k\}})$}

\State Update Critic's parameter
\State \textcolor{gray}{$ \mathbf{w}_{\{t+1,k+1\}} = \mathbf{w}_{\{t,k\}} + lr_{c} \cdot d_{\{t,k\}} \cdot \mathbf{z}_{\{t,k\}}$} 
\State Update the eligibility vector 
\State \textcolor{gray}{$\mathbf{z}_{\{t+1,k+1\}} = \lambda \cdot \mathbf{z}_{\{t,k\}} + \mathbf{s}_{\{t+1,k\}}$}

\State Compute Actor's gradient
\State \textcolor{gray}{$\mathbf{g}_{\{t,k\}} = \frac{d_{\{t,k\}}}{\mathbf{s}_{\{t,k\}}}$}
\State \textcolor{black}{Update Actor's policy parameters}
\State \textcolor{gray}{$ \theta_{\{t,k+1\}} = \texttt{Adam}(\theta_{\{t,k\}}, \mathbf{g}_{\{t,k\}}, lr_{a})$}
\State \Return \textcolor{black}{$\theta_{\{t,k+1\}}$}
\EndFunction

\State

\Function{$\textcolor{comments}{\texttt{Predict}}$}{$agent$}
\State Compute $LP$ based on (13)
\If{$agent=Agent\_PS$}
    \State Compute policy prediction $P_{t}$
    \State \textcolor{gray}{$P_{t} = LP_{t}$}
    \State Compute corresponding $a_{t}$
    \State \textcolor{gray}{$a_{t} = a_{t, k-1} + m \cdot P_{t} \cdot a_{t, k-1}$}
    \State \Return \textcolor{black}{$a_{t}$}
\EndIf

\If{$agent=Agent\_ICR$}
    \State Compute $SP$ based on (14)
    \State Compute policy prediction $P_{t}$
    \State \textcolor{gray}{$P_{t} = \alpha_{LP} \cdot LP_{t} + (1 - \alpha_{LP}) \cdot SP_{t}$}
    \State Compute corresponding $a_{t}$
    \State \textcolor{gray}{$a_{t} = a_{t, k-1} + m \cdot P_{t} \cdot a_{t, k-1}$}
    \State \Return \textcolor{black}{$a_{t}$}
\EndIf
\If{$agent=Agent\_Basal$}
    \State Compute policy prediction $P_{k}$
    \State \textcolor{gray}{$P_{k} = LP_{k}$}
    \State Compute corresponding $a_{k}$
    \State \textcolor{gray}{$a_{k} = a_{k-1} + m \cdot P_{k} \cdot a_{k-1}$}
    \If{$a_{k} < 0.25 \cdot TDD_{k-1}$} 
        \State \textcolor{gray}{ $a_{k} = a_{k-1}$} 
    \EndIf
    \State \Return \textcolor{black}{$a_{k}$}
\EndIf
\EndFunction
\end{algorithmic}
\end{algorithm}

\section{Experimental Setup}

\begin{table}[ht]
    \centering
    \begin{tabular}{|ccc|}
    \toprule
         Characteristics & T1D & T2D \\
         \midrule
        Age, years & Adults & $62(3)$\\
        Female, $\%$ & $50$ & $50$\\
        Body Weight, kg & $69.7(12.4)$ & $95.0(16.5)$\\
        Insulin, U/day/kg & $0.61(0.18)$ & - \\
        CHO ratio,g/U & $15.9(5.3)$ & - \\
        FBG, mg/dL & $119.6(6.7)$ & $154(28)$\\
        OGTT, 2-hour BG, mg/dL & - & $276(57)$\\
        Fasting C-peptide, ng/mL & - & $3.22(1.12)$\\
        \bottomrule
    \end{tabular}
    \caption{Metabolic characteristics of the T1D and T2D in-silico population. Reported mean(SD). Notes: Population glucose measures in the $100$ adult subjects derived from a $75$ gm, $2$ hour Oral Glucose Tolerance test (OGTT) which meet criteria for diabetes diagnosis per clinical consensus standards \cite{sieber2020silico, man2014uva}. FBG is the fasting blood glucose.}
    \label{tab:population}
\end{table}

\subsection{ABBA HyperParameter Setup}\label{ABBA Hyperparameter Setup}

We summarise the hyperparameters of ABBA in \Cref{tab:parameters}. 
The hyperparameters were similar among the virtual subjects. 
Symbol $\rightarrow$ indicates the change of the hyperparameters when the subject is considered to have increased glycaemic variability, or hyperparameters that differ between T1D and T2D; the notation ``/ " is used to show the specific values for each condition. 
We set $\alpha_{SP}$ to $0.1$ after trial and error.
if the person is classified as high risk and experiences increased glycaemic variability within the first two weeks, the initial actor learning rate for $ICR_3$ is set to $0.001$. If the person is at high risk, meaning they frequently experience nocturnal hypoglycaemia or show increased glycaemic variability in the first two weeks, the initial actor learning rate for $ICR_3$ is set to $0.01$. In all cases, we reduce the learning rate, as slower adaptation is safer for individuals with a higher risk of hypoglycaemia.

\begin{table}[ht]
    \centering
    \caption{Hyperparameters of ABBA. $\rightarrow$ indicates the change of the hyperparameters when the subject is considered as an outlier. ``/ " separates the different values between type 1 and 2 diabetes mellitus.}
    \begin{tabular}{|lcc|}
    \toprule
         Hyperparameter & Symbol & Value \\
         \midrule
         Discount factor & $\gamma$ & $0.9$\\
         Eligibility trace decay factor & $\lambda$ & $0.5$ \\
         Critic learning rate for ICR & $lr_{c}$ & $0.1 \rightarrow 0.05$  \\
         Critic learning rate for PS & $lr_{c}$ & $0.1 \rightarrow 0.01$ \\
         Critic learning rate for Basal & $lr_{c}$ & $0.1 \rightarrow 0.01$ \\
         Actor learning rate for ICR & $lr_{a}$ & $0.1 \rightarrow 0.01$\\
         Actor learning rate for PS & $lr_{a}$ & $0.1 \rightarrow 0.01$ \\
         Actor learning rate for Basal & $lr_{a}$ & $0.1 \rightarrow 0.01$ \\
         Weight of $SP$ & $\alpha_{SP}$ & $0.1$\\
         Smoothing factor & m & 0.5 / 1 \\
         \bottomrule
    \end{tabular}
    \label{tab:parameters}
\end{table}

\section{Results}

\begin{table*}[!htp]
    \centering
    \caption{In-silico results show the performance over time of ABBA compared to BBA, with the full cohort over the first and the last $4$ weeks. Data are mean(SD) or median(IQR).
    $\dagger$ $p$-value$< 0.01$ compared to BBA.}
    \resizebox{\textwidth}{!}{%
    \begin{tabular}{|c|c|c|c|c||c|c|c|c|}
        \toprule
         & \multicolumn{4}{c||}{\textbf{First $4$ Weeks}}  &  \multicolumn{4}{|c|}{\textbf{Last $4$ Weeks}} \\
         \midrule
         & TIR, \% & TBR I, \% & TBR II, \% & TAR, \% & TIR, \% & TBR I, \% & TBR II, \% & TAR, \%\\
         \midrule
         \multicolumn{9}{|c|}{\textbf{Adults - T1D}} \\
         \midrule
         \textbf{BBA}& $70.4(14.2)$ & $3.0(0.7-6.2)$ & $0.0(0.0-1.3)$ & $25.2(12.5)$ & $71.7(14.0)$ & $1.4(0.1-4.2)$ & $0.0(0.0-1.8)$ &  $24.9(12.2)$ \\
         \textbf{ABBA}& $72.7(12.5)^{\dagger}$ & $0.6(0.0-1.2)^{\dagger}$ & $0.0(0.0-0.1)^{\dagger}$ & $26.3(11.7)$ & $89.0(79.5-93.1)^{\dagger}$ & $0.4(0.0-1.1)^{\dagger}$ & $0.0(0.0-0.3)^{\dagger}$ & $10.6(6.8-19.9)^{\dagger}$\\
         \midrule
         \multicolumn{9}{|c|}{\textbf{Adults - T2D}} \\
         \midrule
         \textbf{BBA}& $68.22 \pm 19.81$ & $7.79 \pm 12.46$ & $2.86 \pm 6.34$ & $23.98 \pm 21.75$ & $69.67 \pm 20.22$ & $7.47 \pm 12.43$ & $2.71 \pm 6.34$ & $22.85 \pm 22.08$ \\
         \textbf{ABBA}& $75.92 \pm 17.97^{\dagger}$ & $1.61 \pm 2.38^{\dagger}$ & $0.32 \pm 0.78^{\dagger}$ & $22.4 \pm 18.36^{\dagger}$ & $83.66 \pm 16.54^{\dagger}$ & $0.64 \pm 1.26^{\dagger}$ & $0.11 \pm 0.32^{\dagger}$ & $15.70 \pm 16.30^{\dagger}$\\
         \bottomrule
    \end{tabular}}
    \label{tab:res1month}
\end{table*}

\begin{table*}[!htp]
    \centering
    \caption{In-silico results, with the full cohort over the initialisation period, first week after the initialisation, and the last week of simulation. Data are mean(SD).}
    \begin{tabular}{|c|c|c|c||c|c|c||c|c|c|}
        \toprule
         & \multicolumn{3}{c||}{\textbf{Week 1 \& Week 2} (Initialisation period)}  &  \multicolumn{3}{|c|}{\textbf{Week 3}} &  \multicolumn{3}{|c|}{\textbf{Week 14}}\\
         \midrule
         & TIR, \% & TBR, \% & TAR, \% & TIR, \% & TBR, \% & TAR, \% & TIR, \% & TBR, \% & TAR, \% \\
         \midrule
         \textbf{T1D}& $71.7(13.0)$ & $3.6(3.1)$ & $24.7(12.1)$ & $69.7(14.2)$ & $2.3(3.2)$ & $28.0(12.8)$ & $86.7(11.8)$ & $0.3(0.3)$ & $13.0(11.5)$ \\
         \midrule
         \textbf{T2D}& $63.5(21.5)$ & $2.6(6.3)$ & $33.9(22.5)$ & $70.8(19.4)$ & $4.8(7.7)$ & $24.4(20.6)$ & $84.1(16.7)$ & $0.4(1.3)$ & $15.5(16.5)$\\
         \bottomrule
    \end{tabular}
    \label{tab:diffperiods}
\end{table*}

\begin{figure}[htb]
    \centering
    \includegraphics[width=0.9\linewidth]{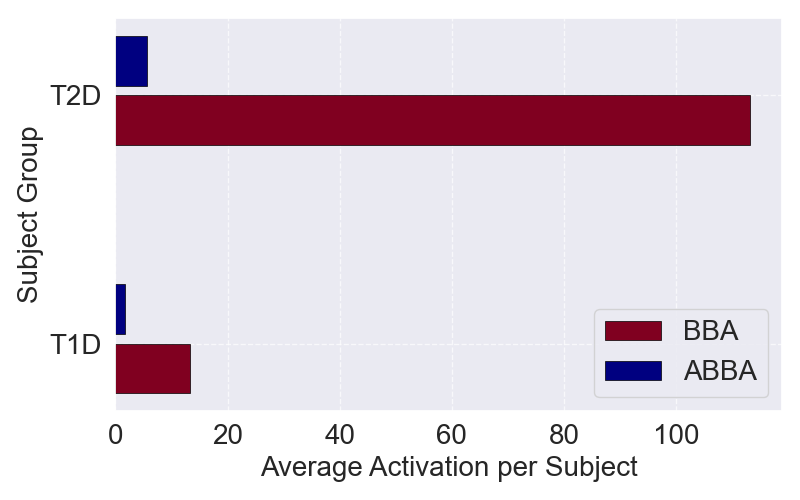}
    \caption{Activation of the rescue meal controller. ABBA requires considerably fewer rescue meal activations compared to BBA, implying better BG control.}
    \label{fig:rescue_meal}
\end{figure}

\Cref{fig:rescue_meal}, shows the activation of the rescue meal controller.
Throughout the entire stimulation the rescue meal controller for BBA was activated in $12$ T1D subjects with the frequency ranging from $1$ to $36$ times (mean $13.33$ times $(10.57)$ per subject). 
For T2D, the rescue meal was activated in $13$ subjects with frequency ranging from $5$ to $528$ (mean $113.23$ times $(175.05)$ per subject). 
For ABBA, the rescue meal controller was activated in $19$ T1D subjects with frequency ranging from $1$ to $7$ times (mean $1.79$ times $(1.55)$ per subject), and in $10$ T2D subjects with the frequency ranging from $1$ to $11$ times (mean $5.60$ times $(3.72)$ per subject).

Upon reviewing the results, it becomes evident that the activation of the rescue meal controller occurred less frequently when utilising ABBA compared to the BBA. 
However, it is important to note that among individuals with T2D the initial parameters of ICR, CF, and basal insulin are sub-optimal.
For instance, a T2D subject under BBA experienced $528$ hypoglycaemic events.
This discrepancy in parameter optimisation arises from our simulator settings, where initial parameters for people with T2D were derived using equations outlined in Section Baseline Method of the main paper, leading to their non-optimality, given partially imprecise initial values the algorithm required a longer period to converge.
In contrast, for people with T1D, the simulator directly provided these values, avoiding extreme behaviours. 
In real-life scenario's, these parameters would typically be provided by clinicians.

\begin{figure}[htb]
    \centering
    \includegraphics[width=0.9\linewidth]{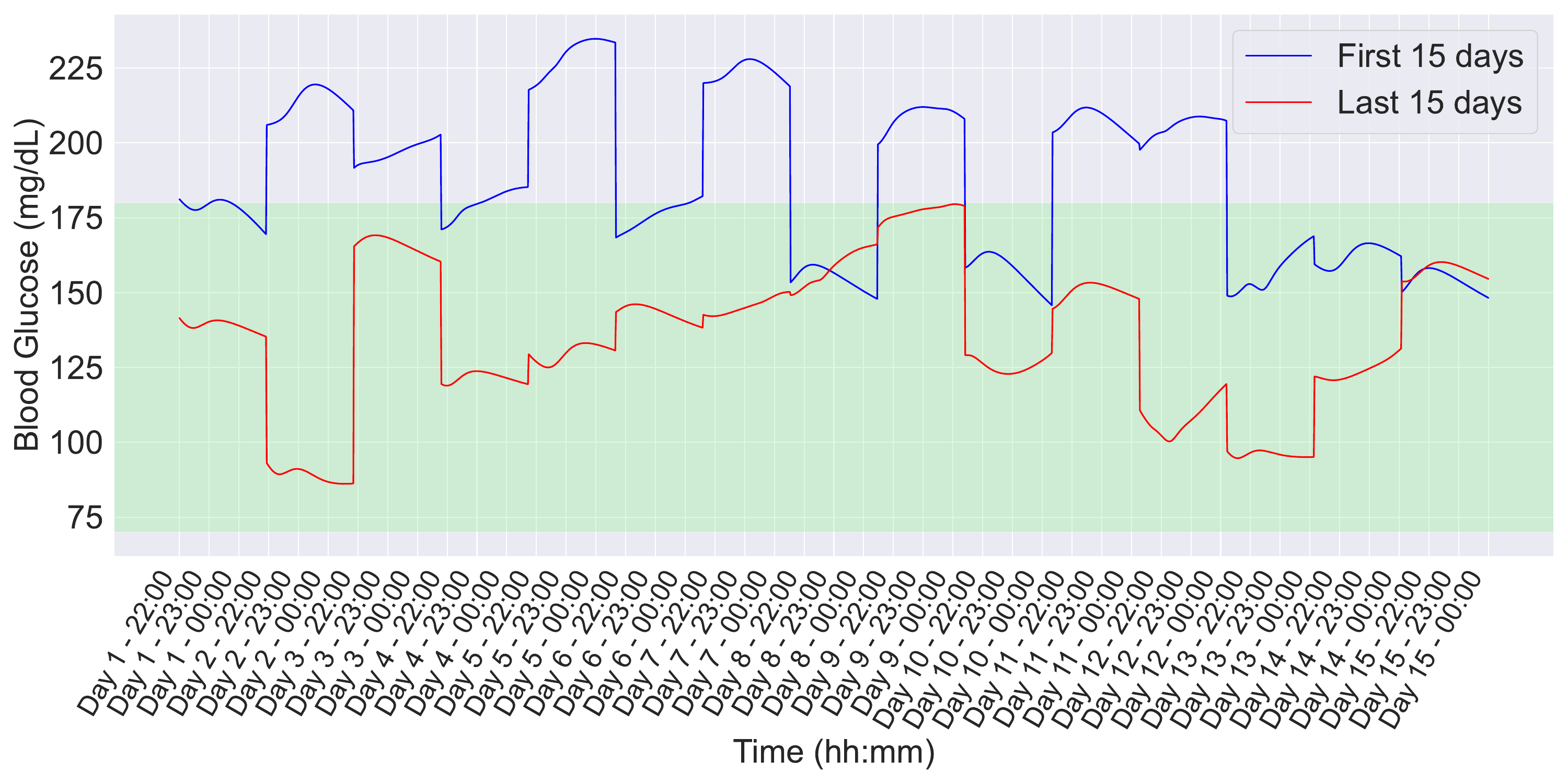}
    \caption{The blood glucose profile for a random adult from 22:00 to 00:00 for the first and the last 15 days. The person has 53\% TAR in the first 15 days of our simulation. However,
    as ABBA updated its parameters, this percentage dropped
    to 0\%.}
    \label{fig:random_adult}
\end{figure}

\subsection{Different Scenarios}
The following scenarios are considered:
\begin{itemize}
    \item Scenario 1: As described in Section Experimental Scenario (main paper).
    \item Scenario 2: Scenario 1 with an additional $\pm 30\%$ inter-day variability of insulin sensitivity \cite{zhu2020basal, sun2018dual}.
    \item Scenario 3: Scenario 1 with the only change of introducing meal misestimation up to $\pm 50\%$.
    \item Scenario 4: Scenario 1 with additional correction boluses. To this end, the person with diabetes also recorded the post-prandial blood glucose measurement (2 hours after the meal). The correction boluses are based on the three different PS factors using \Cref{eq:cb}.
    \begin{equation}
    \label{eq:cb}
    \resizebox{0.8\hsize}{!}{$
        CorrectionBoluses = (\frac{G_c - G_t}{CF}) \cdot PS - IOB, \text{if } G_c > 180 mg/dL
    $}
    \end{equation}
    where $G_c$: current glucose, $G_t$: target glucose, $CF$: correction factor, $PS$: physiological state, and $IOB$: insulin on board.
\end{itemize}

\begin{table*}[htb]
    \centering
    \resizebox{\textwidth}{!}{%
    \begin{tabular}{|c|c|c|c|c|c|c|c|c|c|c|}
        \toprule
        \textbf{T1D} & \textbf{Method} & \textbf{TIR \%} & \textbf{TBR I, \%} & \textbf{TBR II, \%} & \textbf{TAR, \%} & \textbf{\# Hypo. events} & \textbf{\# Hyper. events} & \textbf{Mean BG} & \textbf{Max. BG} & \textbf{Min. BG}\\
         \midrule
         {Scenario 1} & {BBA} & $71.7(13.4)$ & $2.7(3.0)$ & $0.6(1.9)$ & $25.6(11.8)$ & $26.0(24.6)$ & $195.5(42.9)$ & $149(13)$ & $253(41)$ & $55(14)$ \\
         {Scenario 1} & {ABBA}& $83.2(7.7)$ & $0.3(0.3)$ & $0.06(0.08)$ & $16.4(7.6)$ & $5.6(5.4)$ & $195.8(50.7)$ & $148(7)$ & $250(28)$ & $52(19)$\\
         \midrule
         {Scenario 2} & {BBA}& $63.4(14.5)$ & $3.5(2.7)$ & $0.7(1.7)$ & $33.1(13.8)$ & $31.3(22.1)$ & $201.6(29.9)$ & $156(17)$ & $267(41)$ & $45(13)$ \\
         {Scenario 2} & {ABBA}& $78.6(9.0)$ & $0.9(0.8)$ & $0.2(0.3)$ & $20.6(8.9)$ & $10.5(8.4)$ & $199.4(35.3)$ & $149(10)$ & $264(27)$ & $45(17)$\\
         \midrule
         {Scenario 3} & {BBA}& $68.6(10.3)$ & $8.9(4.1)$ & $1.1(2.1)$ & $22.6(10.3)$ & $72.8(29.0)$ & $174.9(44.5)$ & $141(13)$ & $261(44)$ & $41(9)$ \\
         {Scenario 3} & {ABBA}& $77.7(8.0)$ & $2.0(1.5)$ & $0.7(0.5)$ & $20.2(7.6)$ & $22.7(14.4)$ & $180.9(38.5)$ & $147(9)$ & $256(34)$ & $36(12)$\\
         \midrule
         {Scenario 4} & {BBA}& $67.7(13.2)$ & $3.5(2.7)$ & $0.3(0.6)$ & $28.7(12.8)$ & $38.3(41.3)$ & $200.9(41.3)$ & $151(14)$ & $260(39)$ & $50(14)$ \\
         {Scenario 4} & {ABBA}& $87.2(7.4)$ & $0.9(0.6)$ & $0.1(0.1)$ & $11.9(7.1)$ & $12.7(9.9)$ & $177.8(49.6)$ & $142(7)$ & $247(35)$ & $47(14)$\\
         \midrule
         \midrule
         \textbf{T2D} & \textbf{Method} & \textbf{TIR \%} & \textbf{TBR I, \%} & \textbf{TBR II, \%} & \textbf{TAR, \%} & \textbf{\# Hypo. events} & \textbf{\# Hyper. events} & \textbf{Mean BG} & \textbf{Max. BG} & \textbf{Min. BG}\\
         \midrule
         {Scenario 1} & {BBA}& $64.9(17.1)$ & $5.8(8.6)$ & $0.7(1.4)$ & $29.3(21.2)$ & $34.9(49.3)$ & $197.9(95.8)$ & $150(36)$ & $304(71)$ & $75(30)$ \\
         {Scenario 1} & {ABBA}& $76.8(13.0)$ & $1.2(2.5)$ & $0.2(0.6)$ & $22.0(13.4)$ & $9.9(18.8)$ & $222.1(74.4)$ & $144(20)$ & $300(61)$ & $68(21)$\\
        \midrule
         {Scenario 2} & {BBA}& $62.4(17.8)$ & $4.9(7.7)$ & $1.4(2.6)$ & $32.7(22.0)$ & $30.6(46.5)$ & $200.0(97.9)$ & $155(39)$ & $316(73)$ & $75(31)$ \\
         {Scenario 2} & {ABBA}& $71.6(16.1)$ & $1.2(1.8)$ & $0.3(0.5)$ & $27.2(16.6)$ & $11.4(14.4)$ & $225.5(77.2)$ & $148(24)$ & $305(62)$ & $60(23)$\\
        \midrule
         {Scenario 3} & {BBA}& $63.8(16.7)$ & $7.7(10.5)$ & $1.4(2.6)$ & $28.5(21.1)$ & $49.0(64.3)$ & $28.5(21.1)$ & $147(36)$ & $297(69)$ & $70(30)$ \\
         {Scenario 3} & {ABBA}& $75.5(13.5)$ & $1.5(1.9)$ & $0.3(0.5)$ & $23.0(14.2)$ & $16.8(19.4)$ & $226.9(77.4)$ & $143(21)$ & $289(61)$ & $58(21)$\\
         \midrule
         {Scenario 4} & {BBA}& $65.1(17.1)$ & $5.2(8.3)$ & $0.6(1.2)$ & $29.7(21.0)$ & $30.9(46.9)$ & $197.7(95.3)$ & $151(36)$ & $307(71)$ & $75(31)$ \\
         {Scenario 4} & {ABBA}& $77.7(12.4)$ & $1.4(1.9)$ & $0.1(0.3)$ & $21.0(12.7)$ & $14.6(17.7)$ & $224.3(76.1)$ & $142(20)$ & $294(59)$ & $62(23)$\\
         \bottomrule
    \end{tabular}}
    \caption{Results for 11 adults with T1D and 11 adults with T2D with the DMMS.R simulator across different experimental scenarios, comparing ABBA and BBA. Data are mean(SD). The mean, maximum, and, minimum glucose values are expressed in mg/dL.}
    \label{tab:diffScenarios}
\end{table*}

\Cref{tab:diffScenarios} presents the results of Scenarios 1, 2, 3 and 4. 
In all scenarios, ABBA outperformed BBA in terms of TIR while also reducing TBR and TAR. 
Furthermore, over time, ABBA demonstrated continuous improvement across all scenarios.

In the first four weeks, ABBA showed increases in TIR for people with T1D by $3.3 \pm 4.4\%$, $2.9 \pm 5.7\%$, $-1.9 \pm 4.0\%$, and $13.0 \pm 9.5$ along with reductions in TBR by $2.6 \pm 2.7\%$, $2.2 \pm 2.5\%$, $3.3 \pm 3.6\%$, and  $1.9 \pm 2.9$ and TAR by $0.8 \pm 4.5\%$, $0.7 \pm 7.1\%$, $-5.2 \pm 4.2\%$, and $11.1 \pm 10.6$ for Scenarios 1, 2, 3 and 4 respectively, compared to BBA. 
In the last four weeks, ABBA resulted in a significantly higher TIR of $16.4 \pm 10.9\%$, $23.9 \pm 10.3\%$, $9.9 \pm 7.8\%$, and $21.6 \pm 11.6$ and increased reductions in TBR and TAR by $1.2 \pm 2.8\%$, $2.6 \pm 3.5\%$, $0.1 \pm 4.2\%$, and $3.1 \pm 2.6$ and $15.2 \pm 10.5\%$, $21.3 \pm 11.8\%$, $9.8 \pm 7.3\%$, and $18.5 \pm 11.5$ for Scenarios 1, 2, 3 and 4, respectively, relative to BBA. 
After two months, ABBA showed a significant increase of $15.1 \pm 8.0\%$, $20.4 \pm 6.5\%$, $11.4 \pm 7.6\%$, and $9.6 \pm 4.9$ in TIR and a decreased of $0.25 \pm 0.7\%$, $0.8 \pm 1.0\%$, $1.3 \pm 2.1\%$, and $0.9 \pm 0.8$ and $15.3 \pm 8.3\%$, $21.2 \pm 6.6\%$, $12.7 \pm 7.3\%$, and $8.7 \pm 4.8$ in TBR and TAR, respectively, for Scenario, 1, 2, 3, and 4, accordingly, in people with T1D.

For individuals with T2D, during the first four weeks, ABBA demonstrated a significant increase in TIR in people with T2D by $7.7 \pm 7.0\%$, $5.8 \pm 5.4\%$, $8.2 \pm 7.2\%$, and $8.4 \pm 6.7$ and a reduction in TBR by $4.1 \pm 7.0\%$, $3.7 \pm 6.0\%$, $6.1 \pm 8.2\%$, and $3.1 \pm 6.9$ and TAR by $3.6 \pm 6.3\%$, $2.1 \pm 4.2\%$, $2.2 \pm 6.0\%$, and $5.4 \pm 6.5$ compared to BBA, for Scenarios 1, 2, 3, and 4, respectively. 
Accordingly, in the final four weeks, ABBA demonstrated a significant increase in TIR in people with T2D by $13.5 \pm 9.1\%$, $11.0 \pm 6.4\%$, $13.5 \pm 8.1\%$, and $14.4 \pm 8.6$ and a reduction in TBR by $3.8 \pm 7.8\%$, $3.4 \pm 7.4\%$, $5.8 \pm 9.5\%$, and $3.8 \pm 7.3$ and TAR by $9.7 \pm 10.5\%$, $7.6 \pm 7.1\%$, $7.7 \pm 8.4\%$, and $10.5 \pm 9.7$, compared to BBA, for Scenarios 1, 2, 3, and 4, respectively. 
After two months, ABBA showed a significant increase of $7.8 \pm 5.6\%$, $4.0 \pm 3.6\%$, $3.7 \pm 4.1\%$, and $7.6 \pm 3.0$ in TIR and a decreased of $0.5 \pm 2.2\%$, $0.7 \pm 2.2\%$, $0.9 \pm 2.7\%$, and $ 1.2 \pm 2.0$ and $7.4 \pm 6.3\%$, $4.8 \pm 3.7\%$, $4.6 \pm 3.8\%$, and $6.4 \pm 3.5$ in TBR and TAR, respectively, for Scenarios 1, 2, 3, and 4, accordingly, in people with T2D.

Therefore, in all the scenarios tested, ABBA consistently outperformed standard BBA, demonstrating its ability to learn and adapt over time to individual needs.

\subsection{Initialisation Period}

\begin{figure}[!hb]
    \centering
    \includegraphics[width=0.8\linewidth]{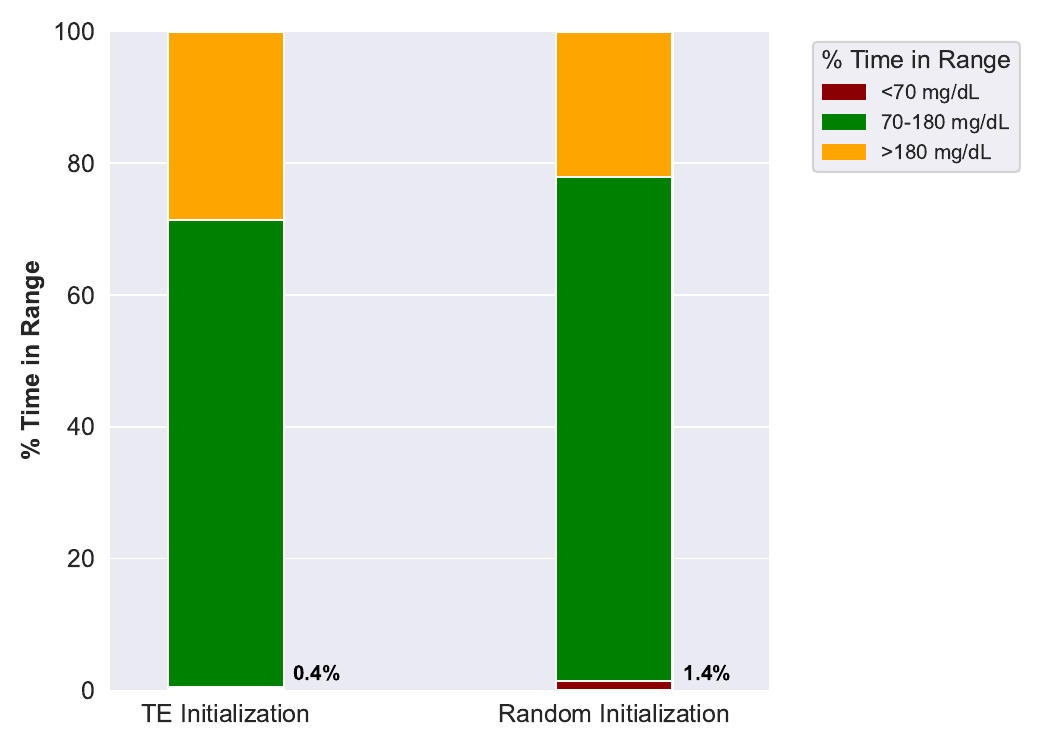}
    \caption{Time in ranges for T1D in the first 2 weeks with TE initialisation and random initialisation.}
    \label{fig:initT1D}
\end{figure}

\begin{figure}[!bt]
    \centering
    \includegraphics[width=0.8\linewidth]{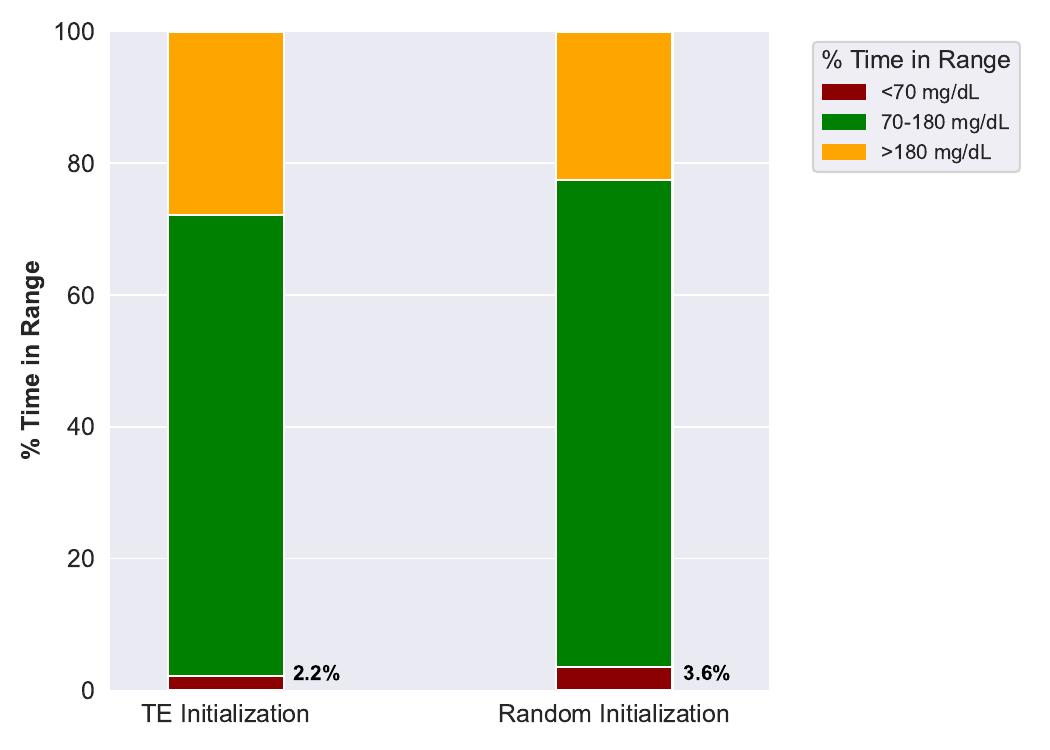}
    \caption{Time in ranges for T2D in the first 2 weeks with TE initialisation and random initialisation.}
    \label{fig:initT2D}
\end{figure}

The primary challenge during the initial phase of algorithm adaptation was the risk of hypoglycaemia. 
To address this, we followed the approach of Daskalaki et al. \cite{daskalaki2013personalized}, where higher TE is associated with smaller changes in the insulin scheme. 
Consequently, the initial values of the policy parameter vectors are set inversely proportional to the estimated TE for each individual. 
By following the method of Daskalaki et al. \cite{daskalaki2013personalized}, we prioritised improving TBR, as shown in \Cref{fig:initT1D} and \Cref{fig:initT2D}, while making slight trade-offs in TIR and TAR during the first two weeks. Compared to random initialization, this approach initially results in a lower TBR, effectively reducing the risk of hypoglycaemia at the beginning of adaptation.

\section*{Acknowledgment}
We acknowledge the support of the Patient Advisory Committee (PAC), who have
provided insights from the perspective of people with diabetes.

\bibliography{IEEEaccess}
\bibliographystyle{plain}

\EOD

\end{document}